\documentclass[akbc,twoside,11pt,lettersize]{article}
\usepackage{akbc}
\usepackage{times}
\usepackage{latexsym}
\usepackage{tabularx}
\usepackage{graphicx}
\usepackage{multirow}
\usepackage{booktabs}
\usepackage{amsmath}
\usepackage{float}
\usepackage{xcolor}
\usepackage{amssymb}% http://ctan.org/pkg/amssymb
\usepackage{pifont}% http://ctan.org/pkg/pifont
\usepackage{enumitem}
\setlist[itemize]{leftmargin=*,label=\scalebox{.8}{\textbullet}}
\usepackage{soul}
\usepackage{subcaption}
\usepackage{cancel}
\usepackage{wrapfig}

\definecolor{greenrgb}{rgb}{0.18, 0.71, 0.18}
\definecolor{lightblue1}{rgb}{0.6, 0.81, 0.93}
\definecolor{lightblue2}{rgb}{0.45, 0.76, 0.98}
\definecolor{lightblue3}{rgb}{0.35, 0.76, 0.98}
\definecolor{purple_corefi}{rgb}{199,21,133}
\definecolor{blue_corefi}{RGB}{45,156,219}
\definecolor{chamoisee}{rgb}{0.63, 0.47, 0.35}
\definecolor{paper_title}{rgb}{0.1, 0.27, 0.58}

\newcommand\data{\textsc{SciCo}}

\akbcheading
\ShortHeadings{\data{}: Hierarchical Cross-Document Coreference for Scientific Concepts}{A. Cattan, S. Johnson, D. Weld, I. Dagan, I. Beltagy, D. Downey, T. Hope}
\finalcopy % Uncomment for camera-ready version, but NOT for submission.
\begin{document}

\title{\data{}: Hierarchical Cross-Document Coreference \\for Scientific Concepts}

\newcommand{\biu}{$^1$}
\newcommand{\allen}{$^2$}
\newcommand{\uw}{$^3$}
\newcommand{\allenuw}{$^{2, 3}$}
\newcommand{\internship}{$^*$}

\author{\name Arie Cattan\biu\internship \email arie.cattan@gmail.com  \\
        \name Sophie Johnson\allen \email sophiej@allenai.org \\
        \name Daniel Weld\allenuw \email danw@allenai.org \\
        \name Ido Dagan\biu \email dagan@cs.biu.ac.il \\
        \name Iz Beltagy\allen \email beltagy@allenai.org \\
        \name Doug Downey\allen \email  dougd@allenai.org \\
        \name Tom Hope\allenuw \email tomh@allenai.org \\
       \addr \biu{}Computer Science Department, Bar Ilan University, Ramat-Gan, Israel \\
       \addr \allen{}Allen Institute for Artificial Intelligence \\ 
       \addr \uw{}Paul G. Allen School for Computer Science \& Engineering, University of Washington\\
      \addr \internship{}Work done during an internship at AI2.
       }

% For research notes, remove the comment character in the line below.
% \researchnote

\maketitle

\begin{abstract}

Determining coreference of concept mentions across multiple documents is a fundamental task in natural language understanding. Previous work on cross-document coreference resolution (CDCR) typically considers mentions of events in the news, which seldom involve abstract technical concepts that are prevalent in science and technology. These complex concepts take diverse or ambiguous forms and have many hierarchical levels of granularity (e.g., tasks and subtasks), posing challenges for CDCR. We present a new task of \emph{Hierarchical} CDCR (H-CDCR) with the goal of \emph{jointly} inferring coreference clusters and hierarchy between them. We create \data, an expert-annotated dataset for H-CDCR in scientific papers, 3X larger than the prominent ECB+ resource. We study strong baseline models that we customize for H-CDCR, and highlight challenges for future work.

% \footnote{\url{https://scico.apps.allenai.org/}}

\end{abstract}

\section{Introduction}
\label{sec:intro}

Cross-document coreference resolution (CDCR) identifies and links textual mentions that refer to the same entity or event across multiple documents. This fundamental task has seen much work recently~\cite{choubey-huang-2017-event, kenyon-dean-etal-2018-resolving, barhom-etal-2019-revisiting, Cattan2021CrossdocumentCR, cattan2021eval, Caciularu2021CrossDocumentLM} and can benefit various downstream applications such as multi-hop question answering~\cite{dhingra-etal-2018-neural, wang-etal-2019-multi-hop}, multi-document summarization~\cite{falke-etal-2017-concept}, and discovery of cross-document relations~\cite{hope2017accelerating,hope2020scisight,hope2021mechanisms}.

Existing datasets for CDCR, such as ECB+ \cite{cybulska2014using}, focus on mentions of news events involving concrete entities such as people or places. Abstract technical concepts are largely unexplored despite their prevalence in domains such as science and technology, and can pose significant challenges for CDCR: they often take diverse forms (e.g., \textsl{class-conditional image synthesis} and \textsl{categorical image generation}) or are ambiguous (e.g., \textsl{network architecture} in AI vs. systems research). These complex concepts also have many hierarchical levels of granularity, such as tasks that can be divided into finer-grained subtasks, where reference to a specific concept entails also a reference to the higher-level concept (e.g., \textsl{CRF} entails the  \textsl{sequence tagging} task), unlike events and entities in ECB+ that are treated as non-hierarchical. 

\begin{wrapfigure}{r}{0.4\textwidth}
  \begin{center}
    \includegraphics[width=0.4\textwidth]{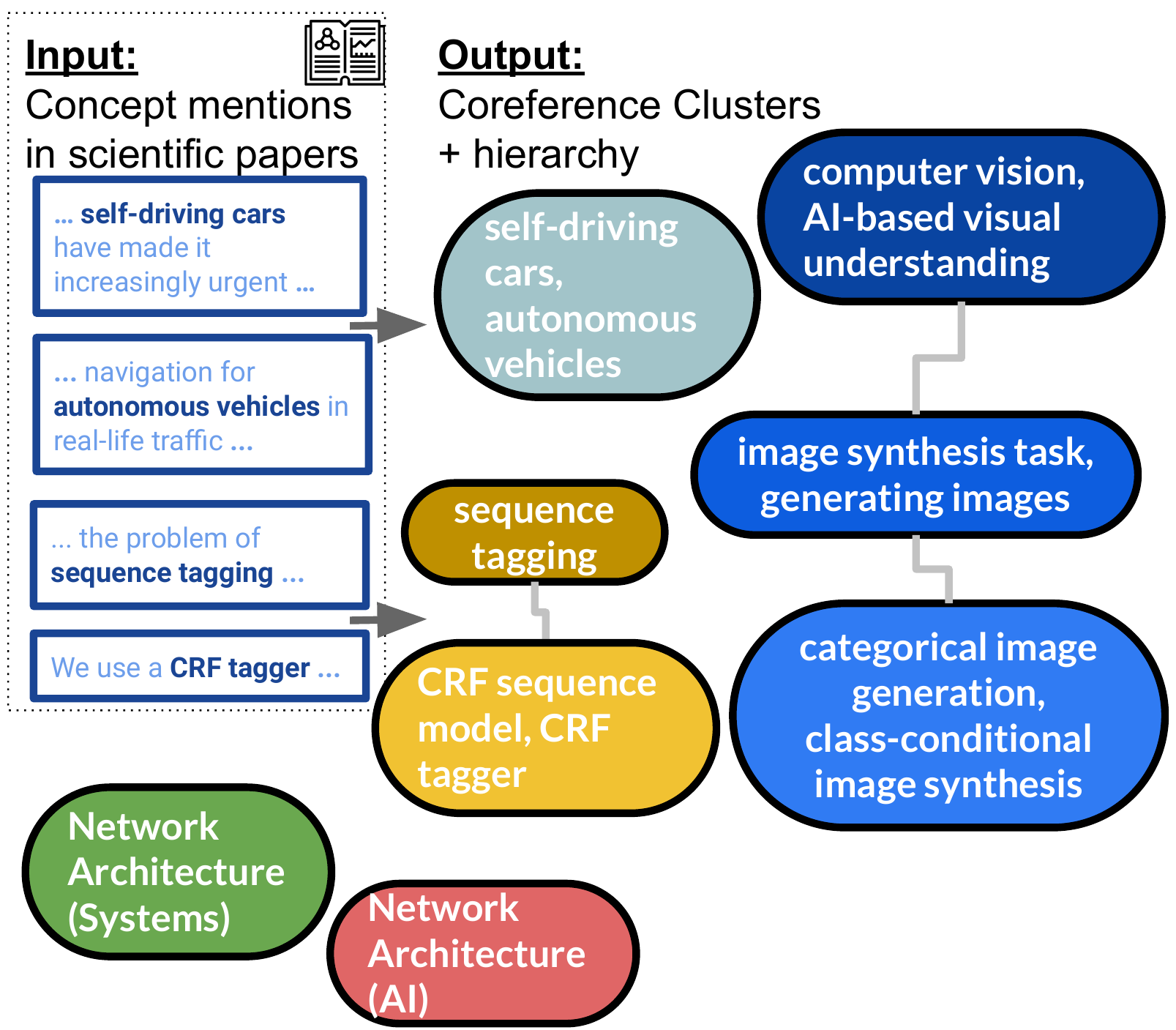}
    \label{fig:teaser}
  \end{center}
  \caption{Given a set of mentions of scientific concepts, our goal is to induce clusters of coreferring mentions and a referential hierarchy between clusters indicating that reference to a child concept (e.g., \textsl{CRF model}) entails reference to a parent concept (\textsl{sequence tagging}).}
  \label{fig:teaser}
\end{wrapfigure}

In this paper, we formulate a novel task of {\em hierarchical CDCR} (H-CDCR). The task is to infer (1) \emph{cross-document entity coreference} clusters of concept mentions in scientific papers, and (2) \emph{referential hierarchy between clusters}, where referring to a child cluster entails reference to the parent. Figure~\ref{fig:teaser} shows the structure we aim to induce given mentions in context. Our task is the first to consider \emph{unified} CDCR and hierarchy between \emph{clusters} of mentions in context, unlike most work focusing on lexicon-level taxonomies over uncontextualized \emph{words} \cite{shwartz2016improving,zhang2018taxogen}.

H-CDCR in science can support many applications. One example illustrating our setting is faceted query by example \cite{mysore2021csfcube}, where given a paper, a user may highlight a specific span (facet) (e.g., \textsl{network architecture}), and explore other papers that refer to the same concept while resolving ambiguity and clustering diverse surface forms. Selecting a facet may also show concept hierarchies enabling users to browse and navigate collections \cite{hope2020scisight} (e.g., by viewing a list of many types of network architectures, automatically identified). See~§\ref{sec:bg} for more discussion of related areas and problems that could benefit from H-CDCR in the scientific domain.

To advance research in this area and enable supervised model training and evaluation, we create a new large-scale dataset named \data\space(\textbf{S}cientific \textbf{C}oncept \textbf{I}nduction \textbf{C}orpus), which is annotated by domain experts.
\data \space consists of clusters of mentions in context and a hierarchy over them, as shown in Figure~\ref{fig:teaser}.  The corpus is drawn from computer science papers, and the concept mentions are {\em methods} and {\em tasks} from across CS.  To build \data{}, we develop a new candidate generation approach %required for large-scale CDCR labelling \cite{Eirew2021WEC}. 
built on three resources: a low-coverage KB, a noisy hypernym extractor, and curated candidates. %Expert annotators use an interface built for our task, achieving high agreement \tom{actually some metrics may not refelect thisX\%}. 
We evaluate strong baseline models, finding that a cross-encoder model addressing coreference and hierarchy jointly outperforms others.\footnote{\data{}, code and models are available at \url{https://scico.apps.allenai.org/}} 

\noindent \textbf{Our main contributions include:} 
\begin{itemize}[topsep=1mm,noitemsep]
    \item We formulate the novel task of hierarchical cross-document coreference (H-CDCR), and explore it within scientific papers. 
    \item We release \data, an expert-annotated dataset 3X larger than the prominent ECB+ CDCR dataset.
    \item We build a model for H-CDCR that outperforms multiple baselines while leaving much room for future improvement.
\end{itemize}

% In the task of cross-document coreference resolution (CDCR) \cite{ barhom-etal-2019-revisiting}, the goal is to cluster mentions of specific people, locations or events.  %\footnote{For example, the OntoNotes \cite{Pradhan2007OntoNotesAU}  guidelines instruct to ignore a mention of \emph{cataract surgery}.}

%%%%%

\section{The H-CDCR Task}
\label{sec:task}

% We start by defining our task and evaluation metrics.

\subsection{Problem Formulation}
\label{subsec:problem}

\begin{table*}[!t]
    \centering
    \resizebox{\textwidth}{!}{
    \begin{tabular}{l p{75mm} p{10mm} p{85mm}}
    \toprule
    \multirow{8}{*}{\textbf{Diversity}} & 
        \textbf{\underline{Title:} \textcolor{paper_title}{Conditional Image Synthesis With Auxiliary Classifier GANs (2017)}}
        && \textbf{\underline{Title:} \textcolor{paper_title}{Mode Seeking Generative Adversarial Networks for Diverse Image Synthesis}} \\
        %  \vspace{2mm}
        &...assessing the discriminability and diversity of \emph{\textbf{\textcolor{purple}{class-conditional image synthesis}}} ... &$\Longleftrightarrow$& 
        ... for \emph{\textbf{\textcolor{purple}{categorical image generation}}}, we apply the proposed method on DCGAN using CIFAR-10 dataset... \\
        \cmidrule{2-4}
        
        & \textbf{\underline{Title:} \textcolor{paper_title}{Detecting anomalous and unknown intrusions against programs}} && \textbf{\textcolor{paper_title}{Fuzziness based semi-supervised learning approach for intrusion detection system}} \\
        
        & ... approaches to \emph{\textbf{\textcolor{purple}{detecting computer security intrusions}}} in real time are misuse detection and... &$\Longrightarrow$& 
        
        \emph{\textbf{\textcolor{purple}{Countering cyber threats}}}, especially attack detection, is a challenging area of research in information... \\
    
        \midrule
    
        \multirow{8}{*}{\textbf{Ambiguity}}
        & \textbf{\underline{Title:} \textcolor{paper_title}{On Generalized and Specialized Spatial Information Grids}}
         && \textbf{\underline{Title:} \textcolor{paper_title}{Query selection for automated corpora construction with a use case in food-drug interactions}} \\
        %  \vspace{2mm}
        &...spatial data acquisition, analysis, \emph{\textbf{{\textcolor{purple}{information extraction}}}} and knowledge discovery ... &$\cancel{\Longleftrightarrow}$& ... building a high-coverage corpus that can be used for \emph{\textbf{{\textcolor{purple}{information extraction}}}} ... \\ 
        
        \cmidrule{2-4}
        
        & \textbf{\underline{Title:} \textcolor{paper_title}{Semi-Supervised Semantic Role Labeling with Cross-View Training (EMNLP 2019)}} && \textbf{\underline{Title:} \textcolor{paper_title}{Learning Mid-level Filters for Person Re-identification (CVPR 2014)}} \\
        & ...SRL model can leverage unlabeled data under the \emph{\textbf{\textcolor{purple}{cross-view training}}}  modeling paradigm...
         &$\cancel{\Longleftrightarrow}$& ... a \emph{\textbf{\textcolor{purple}{cross-view training}}} strategy is proposed to learn filters that are view-invariant and discriminative...\\
        % \midrule

    \bottomrule
    \end{tabular}}
    \caption{\textbf{Examples from \data{}}. Scientific concepts exhibit lexical diversity and ambiguity. For example, \textsl{information extraction} can refer to the NLP task, or to extracting spatial information from grids; and \textsl{cross-view training} can refer to a computer vision technique or a semi-supervised model.}  
    \label{tab:phenomena}
\end{table*}

Our goal is to induce clusters of contextualized mentions that corefer to the same concept, and to infer a hierarchy over these concept clusters. 
Formally, we are given a set of documents $\mathcal{D}$ from a diverse corpus. We assume each $d \in \mathcal{D}$ comes annotated with {\em mentions} (spans of text, see Table~\ref{tab:phenomena}) denoting concepts. Denote by $\mathcal{M}_d = \{m_1, m_2, ..., m_n\}$ the set of mentions in document $d$ and by $\mathcal{M}$ the set of mentions across all $d \in \mathcal{D}$.

The first goal, similar to cross-document coreference resolution, is to cluster the mentions in $\mathcal{M}$ into disjoint clusters $\mathcal{C} = \{C_1, ..., C_t\}$, with each cluster consisting of mentions $\{m | m \in C_i\}$ that all refer to the same underlying concept (see Figure~\ref{fig:teaser}). 
To account for the difficulty in precisely delineating the ``borders'' of extremely fine-grained concepts in scientific literature, clusters are allowed to include subtle variations around a core concept (e.g., \textsl{CRF model}, \textsl{CRF tagger}). 

The second goal is to infer a hierarchy over clusters. Define a graph $G_{\mathcal{C}} = (\mathcal{C},\mathcal{E})$, with vertices representing $\mathcal{C}$ (mention clusters), and directed edges $\epsilon_{ij} \in \mathcal{E}$, each edge representing a hierarchical relation between clusters $C_i$ and $C_j$ which reflects \emph{referential hierarchy}. A relation $\mathcal{C}_1 \xrightarrow[]{} \mathcal{C}_2$ exists when the concept underlying $\mathcal{C}_2$ entails a reference to $\mathcal{C}_1$. For example, a text that mentions the concept ``BERT model'' also (implicitly) invokes several other concepts (``neural language model'', ``neural nets'', ``NLP'') but not others (``robotics'', ``RoBERTa model''). In section~\ref{subsec:anno}, we ground this entailment definition with a faceted search application to assist the annotation.
% (e.g., \textsl{language model} $\xrightarrow[]{}$ \textsl{BERT}).

Put together, the goal in our \textbf{Hierarchical Cross Document Coreference} (H-CDCR) task is: Given documents $\mathcal{D}$ and mentions $\mathcal{M}$, construct clusters $\mathcal{C}$ and a hierarchy graph $G_{\mathcal{C}} = (\mathcal{C},\mathcal{E})$ by learning from a set of $N$ examples $\{(\mathcal{D}^k,\mathcal{M}^k, G^k_{\mathcal{C}})\}^N_{k=1}$. 
In our experiments, we focus on mentions referring to \emph{tasks} and \emph{methods} in computer science papers (see examples in Table \ref{tab:phenomena} and §\ref{sec:scico}). 

\subsection{Evaluation Metrics}
\label{subsec:eval_metrics}

Comparing an extracted set of mention clusters and hierarchies to a gold standard for evaluation is non-trivial. Evaluation metrics for coreference resolution (e.g., MUC, B\textsuperscript{3}, CEAFe, and LEA) do not involve relations between clusters, and models for inferring hierarchical relations such as hypernymy do so over pairs of \emph{individual} terms. Therefore, in addition to reporting established metrics for coreference resolution, we devise two specific metrics for our novel unified task.

\paragraph{Cluster-level Hierarchy Score} 

In H-CDCR, hierarchical relations are defined over \emph{clusters} of mentions. This complicates the evaluation of the hierarchy, since a system may output a different set of clusters from the gold due to coreference errors. 
Our cluster-level score is intended to evaluate hierarchical links without penalizing coreference errors a second time. Let $\mathcal{C}^S_P \xrightarrow[]{} \mathcal{C}^S_C$ be a hierarchical link output by the system between a parent cluster $\mathcal{C}^S_P$ and a child cluster $\mathcal{C}^S_C$. We define this link to be a true positive {\em iff} there exists some pair of mentions in these two clusters that are also in a parent-child relationship in the gold hierarchy. That is, it is positive {\em iff} there exist mentions $p \in \mathcal{C}^S_P$ and $c \in \mathcal{C}^S_C$ which are also in two gold clusters, i.e. $p \in \mathcal{C}^G_P$ and $c \in \mathcal{C}^G_C$, such that $\mathcal{C}^G_P \xrightarrow[]{} \mathcal{C}^G_C$ in the gold.\footnote{We also apply transitive closure, adding edges between pairs of clusters with \emph{indirect} hierarchical relations (e.g., across the first and last nodes in the chain \textsl{computer vision}  $\xrightarrow[]{}$ \textsl{image synthesis}  $\xrightarrow[]{}$ \textsl{categorial image synthesis}), following common practice in hierarchy prediction \cite{li2018smoothing}.} If not, the output link counts as a precision error. We define recall errors analogously, swapping system and gold in the above formalism.

\begin{figure*}[t]
    \centering
    \includegraphics[width=0.7\textwidth]{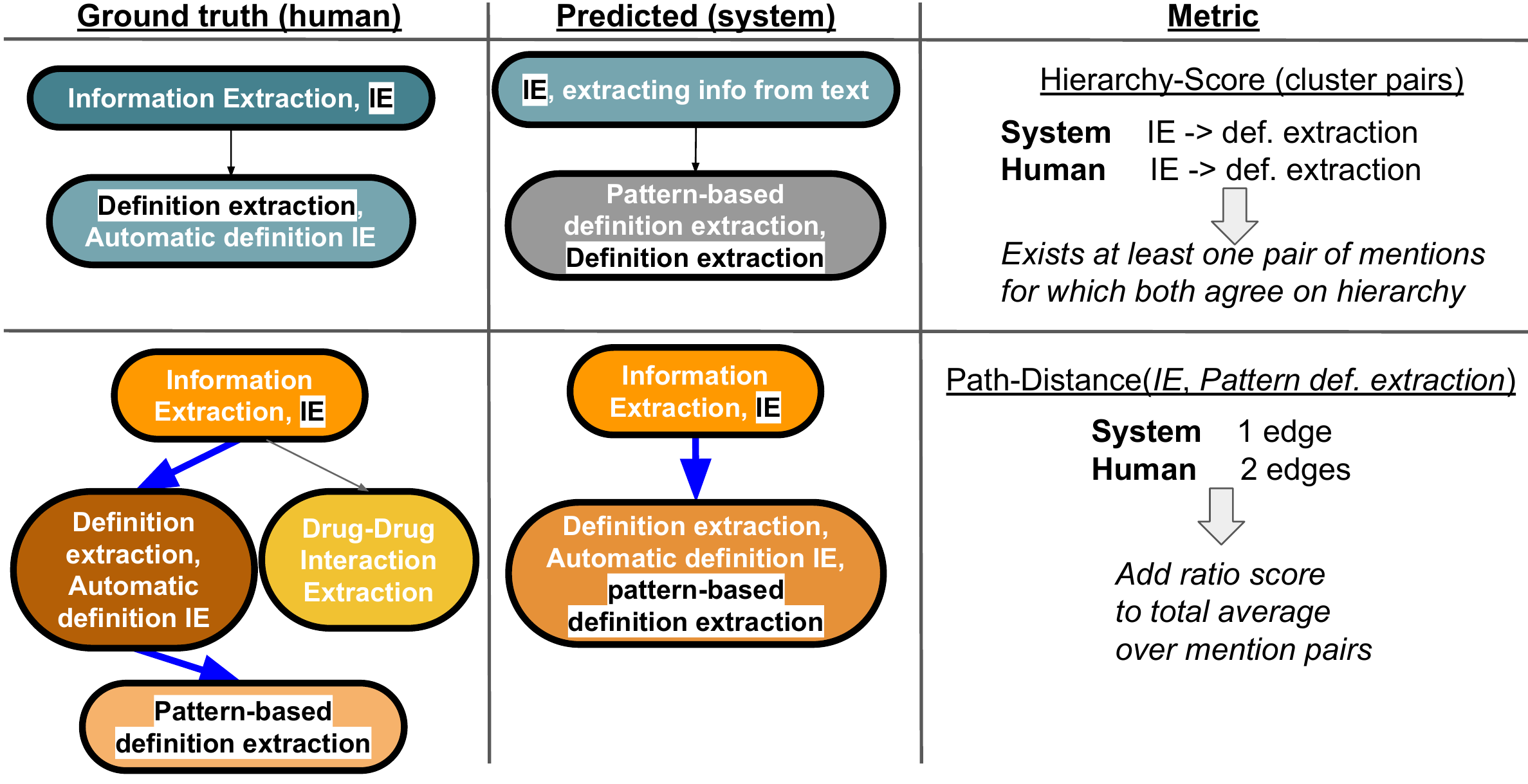}
    \caption{\textbf{Evaluation Metrics Examples}. (Top row) Cluster-level hierarchy score. Both agree that \textsl{IE} is a parent of \textsl{definition extraction}, hence the model is rewarded to avoid doubly penalizing coreference mismatches. (Bottom row) Path-distance score. Graph distances between mention pairs are compared between human and model.}
    \label{fig:path-based}
\end{figure*}

As an example, see Figure~\ref{fig:path-based}, where $p=$ \textsl{IE} and $c=$ \textsl{Pattern-based definition extraction}. %Even though the clusters themselves don't match, 
The metric treats this link as a true positive, since there is at least one pair of mentions in the gold over which the same hierarchy relation holds. Requiring at least one pair, rather than greater overlap, is intended to maximally decouple the coreference and hierarchy penalization. We also report results when requiring a more conservative 50\% overlap, to examine robustness to this choice. 

% \paragraph{Connected Components Agreement} This metric involves segmenting the graph $G_\mathcal{C}$ into connected components, and treating each component as a cluster; we can then compute agreement between two sets of clusterings, one for each annotator, using CoNLL F1. \tom{as illustrated in figure XX} this metric intuitively captures ``coarse-grained agreement'' between annotators, collapsing more granular differences within each respective connected component (such as different sub-clusters, or different hierarchy choices).

\paragraph{Path-Distance Score}

The cluster-level hierarchy score does not take into account the degree of separation between concepts in the graph. We devise a score that does, following previous work on graph-based semantic distances~\citep{Lee1993InformationRB}.

Here, we consider for each pair of mentions $(m_1,m_2)$ the number of cluster-level directed edges needed to traverse from $m_1$ to $m_2$, plus one (mentions in the same cluster have a distance of 1). For each mention pair, we compute this distance for both the ground-truth and the model (see Figure~\ref{fig:path-based}), and the ratio between the smaller of the two distances and the larger. For pairs that are disconnected in the gold or the system but not both, we treat the ratio as zero, and pairs disconnected in both are ignored. The ratios are averaged to obtain the final score. 

More formally, let $W$ be the union of mention pairs that have a path in the gold or system cluster-graph. The path-distance score is defined as follows:

\begin{equation}
    \frac{1}{|W|} \sum_{i, j \in W} \frac{min(p_\text{sys}(i, j), p_\text{gold}(i, j)) + 1}{max(p_\text{sys}(i, j), p_\text{gold}(i, j)) + 1}
\end{equation}
where $p_\text{gold}(i, j)$ and $p_\text{sys}(i, j)$ are the distances between the mentions $i$ and $j$ in the gold and the system, respectively.

This metric gives partial credit for similar but not exactly matching graphs, e.g., a predicted parent-child relation is not considered a complete error if the gold specifies a coreference relation (the ratio is 1/2). 
This can help in ambiguous cases; e.g., whether \textsl{CRF model} is a parent of \textsl{CRF tagger} or belongs in the same cluster.

\begin{figure*}[t]
    \centering
    \includegraphics[width=0.75\textwidth]{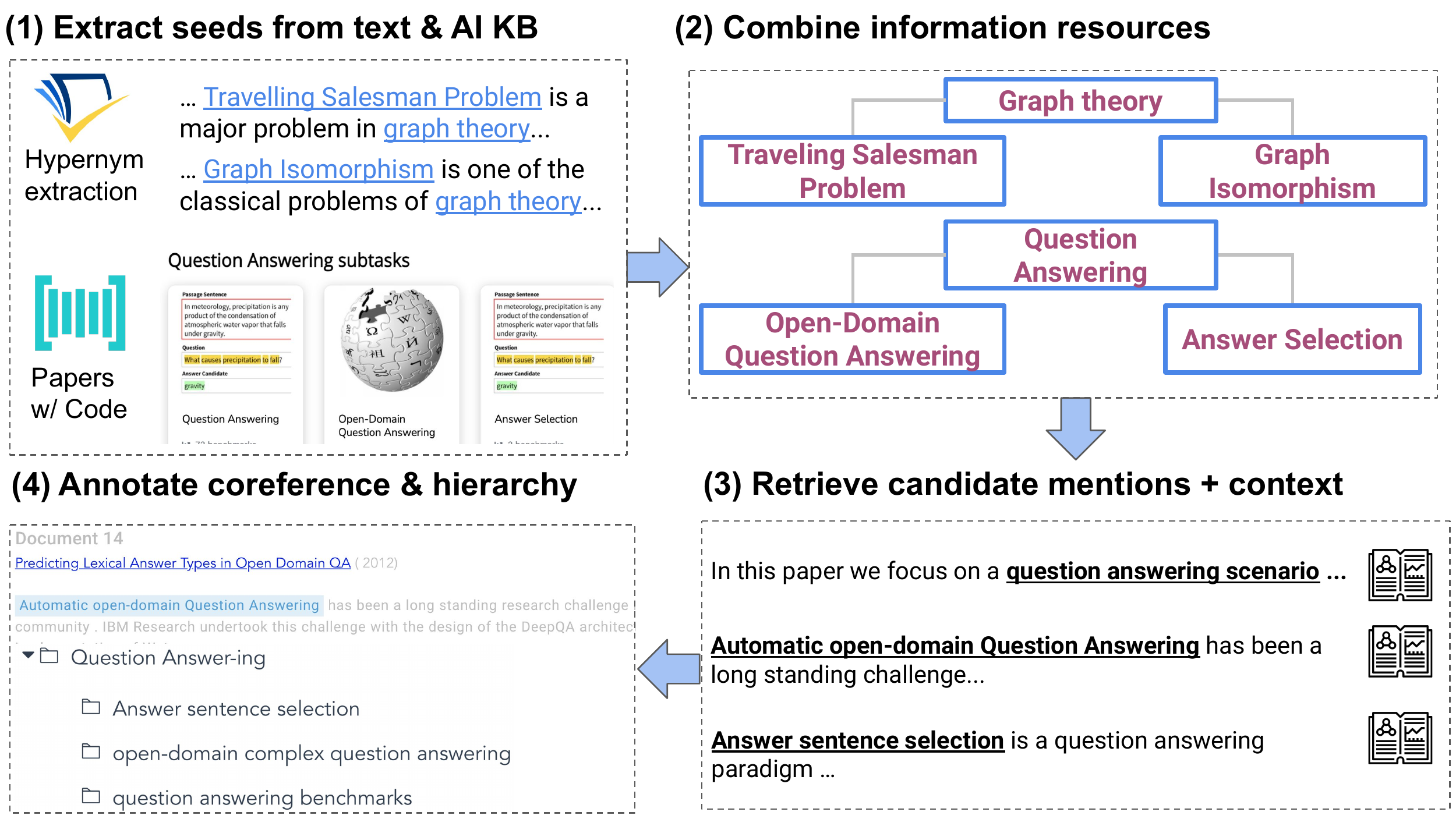}
    \caption{\textbf{Overview of the data collection}. (1) We bootstrap data collection by using PwC and a hypernym extractor. (2) The outputs from these resources are combined, forming groups of interrelated concepts used for candidate generation. (3) We retrieve candidates (mentions in context) based on the constructed concept groups. (4) Expert annotators use an interface built for this task.}
    \label{fig:anno_proc}
\end{figure*}

\section{\data{} Data Construction}
\label{sec:scico}

Obtaining annotations for H-CDCR is non-trivial; showing annotators all possible pairs of documents is not feasible, and presenting randomly drawn mentions is ineffective as they will rarely be related. We thus follow work on data collection for coreference tasks ~\citep{cybulska2014using, jain-etal-2020-scirex, Ravenscroft2021CD2CRCR, Eirew2021WEC} and bootstrap with existing resources.

% As described in ~§\ref{subsec:problem} and illustrated in Figure~\ref{fig:teaser}, the input to our hierarchical CDCR task is a corpus of mentions in context, and the output is a segmentation of these mentions into clusters that corefer to the same concept, with a referential hierarchy between them. \dd{Note, not sure we need to restate this.} \arie{Agreed, what about removing the full paragraph?} \dd{sounds good to me} In this section, we describe how we compile \data{}, a manually-annotated data set for our task that can be used for model training and evaluation.

%The data structure described in ~§\ref{subsec:problem} and illustrated in Figure~\ref{fig:teaser} is a corpus of mentions in context, segmented into clusters that corefer to the same concept, with a referential hierarchy between them. In this section, we proceed to collect data in this form, that can be used for model training and evaluation. 
Our process relies on two primary ingredients: a large corpus of mentions in papers, and a set of seed taxonomies that we leverage to find mentions that are likely to be coreferring or have hierarchical relations. Annotators are then asked to build clusters of mentions and hierarchical relations between them (as in Figure~\ref{fig:teaser}). Below, we describe these two ingredients and how we use them for candidate selection. Our overall data construction approach is illustrated in Figure~\ref{fig:anno_proc}.

%We collect annotations for our task over texts taken from a corpus spanning all computer science \cite{lo-etal-2020-s2orc}. Obtaining useful annotations in our setting is non-trivial; showing annotators all possible pairs of documents is not feasible, and presenting randomly drawn mentions to label is ineffective since random mention pairs will almost never be related. We thus follow recent work on large-scale data collection for coreference resolution ~\citep{singh12:wiki-links, jain-etal-2020-scirex, Ravenscroft2021CD2CRCR, Eirew2021WEC} and bootstrap our annotation process with existing resources.

%Our overall approach is shown in Figure~\ref{fig:anno_proc}. We begin by describing the resources we use. We then describe how we use them for candidate generation.
\subsection{Documents, Mentions and Seed taxonomies}
\label{subsec:docsandmentions}
We populate our dataset of papers and mentions from two data sources: \emph{(i)} a large corpus of 10M CS abstracts from \citep{lo-etal-2020-s2orc} and \emph{(ii)} 438 full-text AI papers from SciREX \cite{jain-etal-2020-scirex}. For the 10M abstracts, we extract mentions referring to methods and tasks, using the DyGIE++ IE model \cite{wadden-etal-2019-entity} trained on SciERC \cite{luan-etal-2018-multi}. SciREX has the advantage of introducing mentions from full paper texts, vetted by a human annotator for quality.

% \subsection{Seed taxonomies} 
% \label{subsec:seed}

We select mention sets from our corpus that are suitable for annotators to label. This requires identifying subsets that are likely to contain coreferent or hierarchically-related mentions. We bootstrap such subsets with existing resources described briefly below (see more details in Appendix~\ref{app:data}).

% \paragraph{High-precision, low-coverage AI KB}
\noindent \textbf{High-precision, low-coverage AI KB.}
Papers With Code\footnote{\url{https://paperswithcode.com}} (PwC) is a public resource that maintains a hierarchical KB of ML related tasks, methods, datasets and results. For example, subtasks of Image Classification include \emph{displaced people recognition} and \emph{satellite image classification}. 

% \paragraph{Corpus-level hypernym extraction}
\noindent \textbf{Corpus-level hypernym extraction.}
To form a higher-coverage resource, we extract a broad (lower-precision) hierarchy of tasks and methods. Specifically, we extract all hypernym relations from the 10M CS abstracts using the DyGIE++ model~\citep{wadden-etal-2019-entity} trained on SciERC. The model extracts hypernym relations between uncanonicalized mentions that appear in the same sentence. We form a hierarchy across the entire corpus by aggregating hyponyms of the same concept string across all papers (see Appendix~\ref{app:hypernym} for technical details). 

\noindent \textbf{Curated list of lexically diverse candidates.}
% \paragraph{Curated list of lexically diverse candidates} 
Automatically collecting lexically diverse mentions is challenging; while our two automated approaches capture such examples, they also capture many lexically \emph{similar} mentions, a common issue in coreference datasets (§\ref{sec:res_analysis}). To increase the diversity of \data{} and enable more transparency in evaluation (§\ref{sec:res_analysis}), we enrich \data{} with a manually-curated collection of 60 groups of closely related but lexically diverse concepts (400 in total). For example, one group contains \{\textsl{deep learning}, \textsl{neural models}, \textsl{DNN algorithms}\}, another includes \{\textsl{class imbalance}, \textsl{skewed label distribution}, \textsl{imbalanced data problem}\}.

%We use a PhD-level annotator with CS and AI background to manually curate a list based on the annotator's knowledge. For example, one curated group contains \{\textsl{deep learning}, \textsl{neural models} \textsl{deep architectures}, \textsl{DNN algorithms}\}, another includes \{\textsl{class imbalance}, \textsl{skewed label distribution}, \textsl{imbalanced data problem}\}, and so forth. We use this list as a further source for candidates.
% \footnote{We also tried a recent scientific paper embedding \cite{cohan-etal-2020-specter} to find topically related documents and random mentions within them, but observed significant noise \dd{I would probably just cut this}.}

\subsection{Candidate Retrieval}
\label{subsec:candidates}

We now turn to how we use the resources described in the previous section in order to select candidate mentions for annotation. Careful candidate generation is often necessary when collecting data for cross-document coreference \cite{cybulska2014using,Ravenscroft2021CD2CRCR}, since the vast majority of random mention pairs are easy negatives. Following standard CDCR terminology, we refer to a batch of candidate mentions to be annotated together as a ``topic''. 
% Annotators are asked to exhaustively label all coreference and hierarchical relationships between pairs of concepts in a given batch.
Given a topic, annotators are asked to form coreference clusters of mentions and label hierarchical relationships between clusters. 
Restricting the number of mentions in a topic is necessary in order to make the quadratic complexity of the task tractable.

We form topics of mentions (Figure~\ref{fig:anno_proc} (3)) as follows. First, we merge the PwC and hypernyms graphs %take the form of directed graphs, with nodes corresponding to concepts and edges to hierarchical entailment relations. 
(de-duplicating nodes) and form groups of candidate concepts where each group consists of a single parent concept and its children. We add our manually-curated groups to this set. We then form a topic from each group by matching each of its concepts against our large-scale corpus (§\ref{subsec:docsandmentions}). Specifically, we retrieve the most similar mentions to each concept, in terms of cosine similarity of embeddings output by a neural model (details in Appendix \ref{app:mention_retrieval}). The union of the retrieved mentions (and their contexts) for a given group form a single topic in our data set (Figure~\ref{fig:anno_proc} (4)). 
The mention retrieval step enriches \data{} with ambiguous cases (e.g., references to \emph{information extraction} with very different meanings), as well as fine-grained variants of concepts.

\subsection{Data Annotation}
\label{subsec:anno}

Dataset annotation for our task is challenging and requires knowledge in computer science research. We hired 4 PhDs and graduate students in CS through UpWork\footnote{\url{http://upwork.com/}}, all authors of at least two scientific publications. Annotators were paid \$20-\$30 per hour (2-3 topics per hour depending on their size). 

In addition to guidelines and tutorials, we also guide annotators to consider the faceted query by example \cite{mysore2021csfcube} application discussed in the Introduction, to help resolve ambiguity. As a concrete example consider two mentions, $m_1=$ \textsl{ELMo model} and $m_2=$ \textsl{ELMo embedding}. Searching for one and retrieving the other should usually be acceptable (indicating coreference). 
By contrast, \textsl{Penn TreeBank POS tagging} should be annotated as a child of \textsl{POS Tagging}, since when issuing a faceted query for the PTB variant of POS tagging, in most cases we would not wish to see a list inundated with variants such as POS tagging in tweets or different languages; conversely, searching for \textsl{POS Tagging} should show a hierarchy of subsumed concepts, including specific variants. While this framing still leaves some room for subjectivity, we embrace it rather than attempt to formulate many complex rules with inevitably limited coverage. To ensure quality we also provided extensive feedback after the first round of annotation.

% while searching for \textsl{POS Tagging} should show \textsl{Penn TreeBank POS tagging} in a hierarchy of results, but not the other direction, indicating referential hierarchy (see Appendix~\ref{app:annotation}). To ensure quality we provided extensive feedback after the first round of annotation.

We provide annotators with a sample of candidate mentions together with their surrounding context and some metadata of the paper such as title, venue, year and the link to the paper itself, as shown in Figure~\ref{fig:anno_proc} (4). To annotate \data{}, we extend CoRefi~\citep{bornstein-etal-2020-corefi}, a recent tool for coreference annotation, by enabling annotation of cluster hierarchies and disaplying metadata~(see Appendix~\ref{app:annotation} for more details). Annotators are asked to annotate both the clusters and relations between the clusters. We repeat this process for each topic in our pool.

\subsection{\data{} Properties}
\label{subsec:scicoprops}

\begin{wraptable}{r}{0.39\textwidth}
    \centering
    \resizebox{0.39\textwidth}{!}{
    \begin{tabular}{@{}lllll@{}}
    \toprule
    & \textbf{Train} & \textbf{Validation} & \textbf{Test} & \textbf{Total} \\
    \midrule
    \# Topics & 221 & 100 & 200 & 521 \\
    \# Documents & 9013 & 4120 & 8237 & 20412 \\
    \# Mentions & 10925 & 4874 & 10423 & 26222 \\
    \# Clusters & 4080 & 1867 & 3711 & 9538\\
    \# Relations & 2514 & 1747 & 2379 & 5981 \\
    %  Width & 7.1 & 6.6 & 6.7 & 6.8 \\
    % Depth & 3.8 & 3.4 & 3.4 & 3.4  \\
    \bottomrule
    \end{tabular}}
    \caption{\data{} statistics.}
    \label{tab:stats}
\end{wraptable}

% \begin{wraptable}{r}{5.5cm}
% \caption{A wrapped table going nicely inside the text.}\label{wrap-tab:1}
% \begin{tabular}{ccc}\\\toprule  
% Header-1 & Header-1 & Header-1 \\\midrule
% 2 &3 & 5\\  \midrule
% 2 &3 & 5\\  \midrule
% 2 &3 & 5\\  \bottomrule
% \end{tabular}
% \end{wraptable} 

Table~\ref{tab:stats} shows summary statistics of \data{}. Notably, \data{} includes over 26K mentions across about 10K clusters and 6K hierarchical relations. \data{} is 3 times larger than prominent CDCR benchmark ECB+~\citep{cybulska2014using}. Mentions in \data{} are taken from over 20K scientific documents  covering diverse CS concepts, larger than ECB+ by an order of magnitude. The average number of connected components across topics is $6.8$, and the average depth of the maximal component (tree) is $3.5$. 

To measure agreement, all annotators labeled the same 40 randomly chosen topics (groups of candidates, totalling about 2200 mentions). Following common evaluation practices in coreference resolution, we measure micro-averaged pairwise agreement (denoted AVG), considering one annotation as gold, and the other as predicted (metrics are symmetric). We also measure the average maximal pairwise IAA, as an ``upper bound'' measure of human performance (MAX-micro). Finally we compute the maximal IAA scores for each topic, and average those 40 scores (MAX-macro).

Models for coreference resolution are traditionally evaluated using different metrics, MUC~\citep{vilain-etal-1995-model}, B\textsuperscript{3}~\citep{bagga-baldwin-1998-entity}, CEAFe~\citep{luo-2005-coreference} and LEA~\citep{moosavi-strube-2016-coreference}, while the main evaluation is CoNLL F1, which is the average F1 of MUC, B\textsuperscript{3} and CEAFe.
For CoNLL F1, AVG is $82.7(\pm2.5)$, MAX-micro is $85.8$ and MAX-macro $90.2$.\footnote{For comparison with OntoNotes~\citep{pradhan-etal-2012-conll}, we also measure the average MUC F1 and report a score of $89.6$, $2.2$ F1 points higher than Ontonotes.} See more coreference metrics in Table~\ref{tab:full_coref_results} in Appendix~\ref{app:coref_results}. For our cluster-hierarchy F1 metric we get AVG of $68.9(\pm2.2)$, MAX-micro of $72.1$ and MAX-macro of $82.3$. The path-distance agreement scores are, respectively, $64.5 (\pm3.5$), $70.0$, and $78.4$. Importantly, as we will see in §\ref{subsec:modelres} these IAA rates are substantially higher than our best model's performance, leaving much room for future modelling.

%. These problems become especially acute when working with a corpus that spans all computer science, with many different research communities and their specialized semantics. This differs from the single document scope, where authors will more often be consistent in terminology. These problems become especially acute when working with a corpus that spans all computer science, with many different research communities and their specialized semantics.

% For models and humans alike, tackling these challenges may often require deep technical understanding of the underlying meaning of concepts (e.g., to understand the strong connection underlying \emph{type I errors} and \emph{false positives}), perhaps using global signals such as publication venues, authors, temporal information, citation links and their contexts
%\input{sections/data_collection}
% \input{sections/04_evaluation}
\section{Models}
\label{sec:model}

We now present our models for H-CDCR.  We start by presenting baselines that separately predict coreference clusters and the hierarchy between them~(§\ref{subsec:baselines}). Then, we describe a  
model that optimizes both subtasks simultaneously~(§\ref{subsec:multiclass}). 

\subsection{Baseline Models}
\label{subsec:baselines}

\paragraph{Coreference Model}
We use a recent state-of-the-art cross-document coreference model~\citep{Cattan2021CrossdocumentCR} for predicting clusters of mentions (denoted by \textbf{CA}; see details in Appendix~\ref{app:cattan_model}). 

We experiment with several variants of this model. First, to evaluate how \data{} differs from existing CDCR datasets, we explore two versions trained only on external CDCR resources: \textbf{CA\textsubscript{News}} trained on ECB+~\citep{cybulska2014using} and \textbf{CA\textsubscript{Sci-News}}, trained on CD\textsuperscript{2}CR~\citep{Ravenscroft2021CD2CRCR} that includes coreference annotation between a single news article and linked scientific paper with mentions extracted using NER. 
Next, we train the model on \data{}. In addition to the RoBERTa (\textbf{CA\textsubscript{RoBERTa}}) pretrained language model used in \cite{Cattan2021CrossdocumentCR}, we explore models specialized to our scientific domain: CS-RoBERTa (\textbf{CA\textsubscript{CS-RoBERTa}})~\citep{gururangan-etal-2020-dont} and SciBERT (\textbf{CA\textsubscript{SciBERT}})~\citep{beltagy-etal-2019-scibert}.

\paragraph{Hierarchy Model}

We consider a referential hierarchical relation between concepts $x$ and $y$ as an \emph{entailment} relation between $y$ and $x$~\citep{glockner-etal-2018-breaking}.\footnote{Indeed, ``John ate an apple" \emph{entails} ``John ate a fruit'' because referring to ``apple'' entails reference to ``fruit".}  
We use a model for textual entailment (RoBERTa-large-MNLI)~\citep{liu2019roberta}, representing each cluster $\mathcal{C}$ by the concatenation of its mentions (without context) \texttt{[CLS]} $\mathcal{C}_k$ \texttt{</s></s>} $\mathcal{C}_j$ \texttt{</s>}, and running RoBERTa-large-MNLI on all $n(n-1)$ cluster pairs $(\mathcal{C}_k, \mathcal{C}_j)$. %considering $\mathcal{C}_k$ as the premise and $\mathcal{C}_j$ as the hypothesis, 
The \texttt{[CLS]} embedding is fed into an output layer for entailment classification. To avoid creation of cyclical graphs, we adopt a greedy approach adding relations iteratively, starting from the highest entailment scores, discarding relations creating cycles. We apply this model to clusters obtained by each baseline.

\subsection{Unified Model for H-CDCR}
\label{subsec:multiclass}

Deciding whether two mentions refer to the same concept or have a hierarchical relation is sometimes non-trivial (e.g \emph{POS Tagging} $\xrightarrow[]{}$ \emph{PTB POS Tagging}, but \emph{artificial neural networks} and \emph{neural networks} refer to the same concept). Therefore, we develop a unified model by formulating our task as multiclass classification, where each mention pair $(m_1, m_2)$ can be assigned into four classes (0) $m_1, m_2$ corefer (1) $m_1$ $\xrightarrow[]{}$ $m_2$ (2) $m_2$ $\xrightarrow[]{}$ $m_1$ or (3) not related. Consider a topic (pool of candidates, ~§\ref{subsec:candidates}), with a set of mentions $\mathcal{M}$. During training, we learn a pairwise scorer $f(\cdot, \cdot)$ by minimizing: 
\begin{equation}
    L = -\frac{1}{N} \sum_{\substack{m_1,m_2 \in \mathcal{M} \\ m_1 \neq m_2}} {y \cdot log(f(m_1, m_2))}
\end{equation}
where $1$-hot $y$ is one of the four classes and $N$ is the number of training pairs. 
For $f(\cdot, \cdot)$ we use Longformer~\citep{Beltagy2020LongformerTL}, a transformer-based language model for processing long sequences so that we can encode pairs of full paragraphs. We also use CDLM, a recent variant of Longformer pre-trained for cross-document tasks~\citep{Caciularu2021CrossDocumentLM}. 

During pretraining, Longformer and CDLM apply \emph{local attention} --- attention only to tokens in a fixed-sized window around each token. When fine-tuning on a specific task, \emph{global attention} --- attention to all tokens in the sequence --- can be assigned to a few target tokens to encode global information.
Following~\citet{Caciularu2021CrossDocumentLM}, we take each mention and its corresponding paragraph, inserting mention markers \texttt{<m>} and \texttt{</m>} surrounding the mention to obtain a mention representation. For CDLM, we add the document markers \texttt{<doc-s>} and \texttt{</doc-s>} surrounding each document. Then, we concatenate the representations of $m_1$ and $m_2$ separated by a separator token \texttt{</s>}, and add a \texttt{[CLS]} token at the beginning. We assign \emph{global attention} to the \texttt{[CLS]} and the mention markers of the two mentions. The \texttt{[CLS]} vector is finally fed into a linear layer $W^{d \times 4}$, for fine-tuning the model.

\begin{wraptable}{r}{0.6\textwidth}
\centering
\begin{subtable}{0.6\textwidth}
\resizebox{\columnwidth}{!}{
\begin{tabular}{lllllll}
    % \begin{table}[]
    % \centering
    % \resizebox{\columnwidth}{!}{
    \toprule
    % & Coreference & \multicolumn{3}{l}{Hierarchy} & CC & Shortest Path Ratio & AUC \\
    % & CoNLL F1 & Recall & Precision & F1 & CoNLL F1 \\
    & \bf{Coreference} & \multicolumn{3}{l}{\bf{Hierarchy}} && \bf{Path} \\
    & CoNLL F1 & F1 & F1-50\% &&& Ratio \\
    \midrule
        IAA (AVG) &  82.7 & 68.9 & 62.8 &&& 64.5 \\
        IAA (MAX-Macro) &  90.2 & 82.3 & 77.7  &&& 78.4\\
    \midrule
         CA\textsubscript{News} &  52.4 & 37.1 & 13.0 &&& 24.1 \\
         CA\textsubscript{Sci-News} &  43.5 & 29.2 & 12.3 &&& 21.6 \\
            %  \midrule
         CA\textsubscript{\data{}} & 55.2 & 23.7 & 15.8 &&& 21.2 \\
         CA\textsubscript{\data{}} + CS-RoBERTa & 57.4 & 23.5 & 16.1 &&& 23.6 \\
         CA\textsubscript{\data{}} + SciBERT & 66.8 & 23.8 & 17.8  &&& 28.4 \\
             \midrule
        %   Pipeline & \bf{77.0} & 38.8 & 29.1 &&& 43.7 \\
        %  Unified\textsubscript{LF-sentence} & 77.0 & 43.0 & 33.9 &&& 45.7 \\
          Unified\textsubscript{Longformer} & \bf{77.2} & 44.5 & \textbf{36.1} &&& \bf{47.2} \\
           Unified\textsubscript{CDLM} & 77.0 & \bf{44.8} & 35.5 &&& 45.9 \\
    \bottomrule
    \end{tabular}}
    % \caption{Caption \tom{add pipeline} \tom{do we need precision/recall here? or appendix?} \tom{adding max iaa here will make it look better} \tom{add macro} \tom{important caption goes here - clarifying why the gap is meaningful, discussing the challenging breakdown when those are added, etc.}}
\end{subtable}
\caption{\textbf{Model results}. We evaluate strong CDCR baselines, and a unified multiclass model for H-CDCR that outperforms the baselines.} 
\label{tab:results}
\end{wraptable}

At inference time, we build clusters using agglomerative clustering over predicted coreference scores, in the same way as in the baseline described above (§\ref{subsec:baselines}). Then, for each pair of clusters $(\mathcal{C}_1, \mathcal{C}_2)$, we aggregate cross-cluster mention-pair predictions for hierarchical relations as follows. Given a pair $(m_1,m_2)$ where $m_1 \in \mathcal{C}_1$ and $m_2 \in \mathcal{C}_2$, we compute the probability score for $m_1$ being a child of $m_2$, and define the score of $C_1$ being the child of $C_2$ as the average of all pairwise scores for all $\{(m_i,m_j) | m_i \in \mathcal{C}_1, m_j \in \mathcal{C}_2\}$: $ s(\mathcal{C}_1, \mathcal{C}_2) = \frac{1}{|\mathcal{C}_1|\cdot|\mathcal{C}_2|} \sum_{m_1 \in \mathcal{C}_1} \sum_{m_2 \in \mathcal{C}_2} f_\text{is-child}(m_1, m_2)$.

To avoid cycles %and multiple nodes linked as direct parent to the same concept
we apply the same greedy approach as in the baseline models~(§\ref{subsec:baselines}) and stop when the hierarchy score is below a tuned threshold. Full implementation details and hyper-parameters are described in Appendix~\ref{app:hp}.

\section{Experimental Results}
\label{subsec:modelres}

Table~\ref{tab:results} presents the results of the baseline models as well as our unified solution. Results of coreference are reported using the CoNLL F1 metric. We also report performance with our two metrics introduced earlier (§\ref{subsec:eval_metrics}). See Appendix~\ref{app:coref_results} for all coreference metrics, as well as recall and precision for the hierarchy score.

The results show that training a model to predict both coreference and hierarchy in a multiclass setup outperforms baselines by a large margin across all metrics.  
In terms of coreference, using the \textbf{CA} baselines (§\ref{subsec:baselines}) trained on external datasests leads to the lowest results. Training the same model on \data{} boosts results, with a significant boost from SciBERT \cite{beltagy-etal-2019-scibert}. 
Focusing on hierarchy, we examine results in terms of our cluster-level hierarchy metric (§\ref{subsec:eval_metrics}) which is designed to not doubly-penalize coreference errors, and the path-based metric. For all baselines, we use a state-of-art pre-trained entailment model \cite{liu2019roberta} (see §\ref{sec:model}) for predicting relations between clusters, which leads to poor results in comparison to the unified model.

\subsection{Ablation Study}
\label{subsec:eval_multi}

\begin{wraptable}{r}{0.5\textwidth}
    \centering
    \resizebox{0.5\textwidth}{!}{
    \begin{tabular}{lllll}
    \toprule
    
    & \bf{Coreference} & \multicolumn{2}{l}{\bf{Hierarchy}} & \bf{Path} \\
    & CoNLL F1 & F1 & F1-50\% & Ratio \\
    \midrule
    
    $-$ unified & 77.1 ($-$0.1) & 41.6 ($-$2.9) & 32.3 ($-$3.8) & 44.2 ($-$3.0)\\
    $-$ paragraph & 77.0 ($-$0.2) & 43.0 ($-$1.5) & 33.9 ($-$2.2) & 45.7 ($-$1.5) \\
    \bottomrule
    \end{tabular}}
    \caption{\textbf{Ablation results}. Parentheses show the relative drop in performance. Both large context and the unified approach contribute to the scores.}
    \label{tab:ablation}
\end{wraptable}
We conduct an ablation study examining our unified formulation and utility of wider contexts. To ablate the unified model (multiclass), we train two models using the same architecture (§\ref{subsec:multiclass}): one for coreference only, the other for hierarchy only. In the former, we consider hierarchical relations as negative pairs, and train a binary classification model with classes (0) unrelated and (1) coreference. The hierarchy-only model has classes (0) unrelated, (1) parent-child and (2) child-parent. During inference, we apply the same approach as in §\ref{subsec:multiclass} for creating clusters and inferring cluster hierarchy.

As shown in Table~\ref{tab:ablation}, the unified model outperforms the disjoint approach in the cluster-hierarchy and path-based scores, indicating the utility of optimizing for both tasks simultaneously. %A possible reason for this gap may be the consideration of both coreference or unrelated pairs in the same class. 
Learning a single model also results in half the number of parameters than in the disjoint approach, while achieving substantially better results. Finally, using only a mention's sentence as context (instead of full paragraph) also leads to a performance drop.

\subsection{Lexical Diversity Impact on Coreference}
\label{sec:res_analysis}

\begin{wraptable}{r}{0.57\textwidth}
\centering
\resizebox{0.47\textwidth}{!}{
\begin{tabular}{lll}
    \toprule
         Test subset & Unified & Edit dist. \\
         \midrule
        Full & 77.2 & 53.3 \\
        Lowest 10\% (edit dist.)  & 64.4 & 27.2 \\
        Lowest 20\% (edit dist.)  & 69.4 & 34.6 \\
        Curated (lexically diverse) & 67.1 & 35.0 \\
        \bottomrule
    \end{tabular}}
\caption{\textbf{CoNLL F1 results by lexical diversity}. We take the bottom 10\% (20\%) topics ranked an edit-distance baseline's CoNLL F1, and also examine manually curated lexically diverse topics (§\ref{subsec:docsandmentions}). Results indicate \data{} contains subsets with varying levels of lexical diversity that correlate with coreference difficulty.} 
\label{tab:editdist}
\end{wraptable}

We explore how coreference performance correlates with lexical diversity (and ambiguity). We examine a simple baseline that uses the Levenshtein edit distance in agglomerative clustering. 
Surface-form matching baselines are known to have comparatively fair performance in CDCR datasets \cite{barhom-etal-2019-revisiting,Eirew2021WEC}. 
We take the bottom 10\% and bottom 20\% of topics ranked by the baseline's CoNLL F1 (20 and 40 topics, respectively). We also examine the set of manually curated lexically diverse topics (§\ref{subsec:docsandmentions}).

In Table~\ref{tab:editdist}, we report micro-averaged results for each subset. Our unified model does substantially worse than its overall performance (e.g., the model does not identify coreference between \textsl{scientific paper analysis} and \textsl{scholarly document analysis}, \textsl{manual annotation} and \textsl{human labeling}).
We also check that the edit-distance baseline does not correlate with inter-annotator agreement, to ensure  IAA rates reported earlier can serve as a measure of human performance across levels of diversity. 
Using Pearson and Spearman correlation tests, we confirm this hypothesis (p-values of 0.10, 0.21).
These results indicate \data{} contains subsets with varying levels of lexical diversity that correlate with coreference difficulty, requiring richer information and models to resolve.

\section{Related Work}
\label{sec:bg}

\paragraph{Entity coreference resolution} 

%The goal in this task is to cluster mentions (text spans) that refer to the same entity. 
Entity coreference work focuses on mentions within a single document (WD), while cross-document (CD) work focuses on coreference between \emph{events} in the news \citep{cybulska2014using, vossen-etal-2018-dont}. In the science domain, some work has been done on entity coreference in the WD setting \cite{luan-etal-2018-multi,jain-etal-2020-scirex}. %Recently, ~\citep{Ravenscroft2021CD2CRCR} built a dataset with pairs of news report and linked scientific papers, with entity mentions automatically extracted. In this dataset corefering mentions are only across single pairs of documents, with pairs known in advance. 
Unlike previous work, in this paper we considered CDCR jointly with inferring hierarchy, for a domain with abstract technical concepts that are nested in many levels of granularity that can be hard to tell apart.

\vspace{-0.15in}

\paragraph{\textbf{Entity linking}}
Entity linking (EL) involves linking mentions of entities to knowledge base (KB) entries. In science, KBs are often scarce and highly incomplete \cite{hope2021mechanisms}. In our work, we used a low-coverage KB to bootstrap data collection and did not assume to have a KB during training or inference. Recent work \cite{Angell2020ClusteringbasedIF} has shown that clustering WD mentions can boost EL in biomedical papers; %by helping resolve generic or specialized forms;
our work on CD mentions could help by pooling information from diverse contexts, and potentially detect entities missing from KBs \cite{lin2012no,wu2016exploring}.

\vspace{-0.15in}

%A less-explored task is that of recognizing mentions of \emph{novel} entities, that are not present in a KB \cite{lin2012no,wu2016exploring}. 
%The ability to automatically discover and ingest novel entities to a KB can be important in rapidly evolving corpora, such as in scientific literature.

\paragraph{\textbf{Taxonomy construction}} Most work in this area generates a graph where nodes are \emph{single} terms representing a concept (e.g., hypernym-hyponym pairs extracted from single sentences) \cite{hearst1992automatic,roller2018hearst}. 
This approach does not resolve ambiguity of identical surface forms (e.g., \emph{information extraction}) and lexical diversity of concepts. Recent work in the data mining community \cite{shang2020nettaxo} focused on unsupervised construction of taxonomies, with clusters of uncontextualized terms given as input. % and using post-hoc human vetting of results \cite{zhang2018taxogen}. 
Our work can potentially help these and related applications \cite{poon2010unsupervised} by introducing context and supervision.

\vspace{-0.15in}

\paragraph{\textbf{Word sense induction}} A related line of work focuses on learning to induce multiple senses of words from text, to capture polysemy and resolve ambiguity. Such work typically employs uncontextualized word embeddings \cite{athiwaratkun2017multimodal, arora2018linear} or phrase-level embeddings \cite{chang2021extending}.
Relatedly, \citet{shwartz-dagan-2016-adding} label pairs of \emph{words} for equivalent senses depending on context, in general language rather than abstract technical concepts. %and only label hierarchical relationships between pairs of \emph{words}, not coreference {\em clusters} as in \data{}.

\vspace{-0.15in}

\section{Conclusion}

We present \data{}, a dataset for a novel task of hierarchical cross-document coreference resolution (H-CDCR) in the challenging setting of scientific texts. \data{} is annotated by domain experts and is three times larger than comparable datasets from the news domain. We evaluate strong baseline models on our data. A joint approach that infers both coreference and hierarchical relationships in the same model outperforms multiple baselines, but leaves substantial room for improvement.

\paragraph{\bf Acknowledgments:} Many thanks to  anonymous reviewers. This project is supported in part by NSF Grant OIA-2033558,  NSF RAPID grant 2040196, and ONR grant N00014-18-1-2193.

%An interesting direction for future work is to explore distant supervision approaches and models that use corpus-level context.

%\input{sections/ethics}

\bibliography{anthology,custom}

\begin{thebibliography}{56}
\providecommand{\natexlab}[1]{#1}
\providecommand{\url}[1]{\texttt{#1}}
\expandafter\ifx\csname urlstyle\endcsname\relax
  \providecommand{\doi}[1]{doi: #1}\else
  \providecommand{\doi}{doi: \begingroup \urlstyle{rm}\Url}\fi

\bibitem[Angell et~al.(2021)Angell, Monath, Mohan, Yadav, and
  McCallum]{Angell2020ClusteringbasedIF}
Rico Angell, Nicholas Monath, Sunil Mohan, Nishant Yadav, and A.~McCallum.
\newblock Clustering-based inference for biomedical {E}ntity {L}inking.
\newblock In \emph{NAACL}, 2021.

\bibitem[Arora et~al.(2018)Arora, Li, Liang, Ma, and Risteski]{arora2018linear}
Sanjeev Arora, Yuanzhi Li, Yingyu Liang, Tengyu Ma, and Andrej Risteski.
\newblock Linear algebraic structure of word senses, with applications to
  polysemy.
\newblock \emph{Transactions of the Association for Computational Linguistics},
  6:\penalty0 483--495, 2018.

\bibitem[Athiwaratkun and Wilson(2017)]{athiwaratkun2017multimodal}
Ben Athiwaratkun and Andrew Wilson.
\newblock Multimodal word distributions.
\newblock In \emph{Proceedings of the 55th Annual Meeting of the Association
  for Computational Linguistics (Volume 1: Long Papers)}, pages 1645--1656,
  2017.

\bibitem[Bagga and Baldwin(1998)]{bagga-baldwin-1998-entity}
Amit Bagga and Breck Baldwin.
\newblock Entity-based cross-document coreferencing using the vector space
  model.
\newblock In \emph{{COLING} 1998 Volume 1: The 17th International Conference on
  Computational Linguistics}, 1998.
\newblock URL \url{https://www.aclweb.org/anthology/C98-1012}.

\bibitem[Barhom et~al.(2019)Barhom, Shwartz, Eirew, Bugert, Reimers, and
  Dagan]{barhom-etal-2019-revisiting}
Shany Barhom, Vered Shwartz, Alon Eirew, Michael Bugert, Nils Reimers, and Ido
  Dagan.
\newblock Revisiting joint modeling of cross-document entity and event
  coreference resolution.
\newblock In \emph{Proceedings of the 57th Annual Meeting of the Association
  for Computational Linguistics}, pages 4179--4189, Florence, Italy, July 2019.
  Association for Computational Linguistics.
\newblock \doi{10.18653/v1/P19-1409}.
\newblock URL \url{https://www.aclweb.org/anthology/P19-1409}.

\bibitem[Beltagy et~al.(2019)Beltagy, Lo, and Cohan]{beltagy-etal-2019-scibert}
Iz~Beltagy, Kyle Lo, and Arman Cohan.
\newblock {S}ci{BERT}: A pretrained language model for scientific text.
\newblock In \emph{Proceedings of the 2019 Conference on Empirical Methods in
  Natural Language Processing and the 9th International Joint Conference on
  Natural Language Processing (EMNLP-IJCNLP)}, pages 3615--3620, Hong Kong,
  China, November 2019. Association for Computational Linguistics.
\newblock \doi{10.18653/v1/D19-1371}.
\newblock URL \url{https://www.aclweb.org/anthology/D19-1371}.

\bibitem[Beltagy et~al.(2020)Beltagy, Peters, and
  Cohan]{Beltagy2020LongformerTL}
Iz~Beltagy, Matthew~E. Peters, and Arman Cohan.
\newblock Longformer: The long-document transformer.
\newblock \emph{ArXiv}, abs/2004.05150, 2020.

\bibitem[Bornstein et~al.(2020)Bornstein, Cattan, and
  Dagan]{bornstein-etal-2020-corefi}
Ari Bornstein, Arie Cattan, and Ido Dagan.
\newblock {C}o{R}efi: A crowd sourcing suite for coreference annotation.
\newblock In \emph{Proceedings of the 2020 Conference on Empirical Methods in
  Natural Language Processing: System Demonstrations}, pages 205--215, Online,
  October 2020. Association for Computational Linguistics.
\newblock \doi{10.18653/v1/2020.emnlp-demos.27}.
\newblock URL \url{https://www.aclweb.org/anthology/2020.emnlp-demos.27}.

\bibitem[Bowman et~al.(2015)Bowman, Angeli, Potts, and
  Manning]{bowman2015large}
Samuel~R Bowman, Gabor Angeli, Christopher Potts, and Christopher~D Manning.
\newblock A large annotated corpus for learning natural language inference.
\newblock In \emph{Conference on Empirical Methods in Natural Language
  Processing, EMNLP 2015}, pages 632--642. Association for Computational
  Linguistics (ACL), 2015.

\bibitem[Caciularu et~al.(2021)Caciularu, Cohan, Beltagy, Peters, Cattan, and
  Dagan]{Caciularu2021CrossDocumentLM}
Avi Caciularu, Arman Cohan, Iz~Beltagy, Matthew~E. Peters, Arie Cattan, and
  I.~Dagan.
\newblock Cross-document {L}anguage {M}odeling.
\newblock \emph{ArXiv}, abs/2101.00406, 2021.

\bibitem[Cattan et~al.(2021{\natexlab{a}})Cattan, Eirew, Stanovsky, Joshi, and
  Dagan]{Cattan2021CrossdocumentCR}
Arie Cattan, Alon Eirew, Gabriel Stanovsky, Mandar Joshi, and Ido Dagan.
\newblock Cross-document coreference resolution over predicted mentions.
\newblock In \emph{Findings of the Association for Computational Linguistics:
  ACL 2021}, Online, August 2021{\natexlab{a}}. Association for Computational
  Linguistics.

\bibitem[Cattan et~al.(2021{\natexlab{b}})Cattan, Eirew, Stanovsky, Joshi, and
  Dagan]{cattan2021eval}
Arie Cattan, Alon Eirew, Gabriel Stanovsky, Mandar Joshi, and Ido Dagan.
\newblock Realistic evaluation principles for cross-document coreference
  resolution.
\newblock In \emph{Proceedings of the Tenth Joint Conference on Lexical and
  Computational Semantics}. Association for Computational Linguistics,
  2021{\natexlab{b}}.

\bibitem[Cer et~al.(2017)Cer, Diab, Agirre, Lopez-Gazpio, and
  Specia]{cer2017semeval}
Daniel Cer, Mona Diab, Eneko Agirre, I{\~n}igo Lopez-Gazpio, and Lucia Specia.
\newblock Semeval-2017 task 1: Semantic textual similarity multilingual and
  crosslingual focused evaluation.
\newblock In \emph{SemEval-2017}, 2017.

\bibitem[Chang et~al.(2021)Chang, Agrawal, and McCallum]{chang2021extending}
Haw-Shiuan Chang, Amol Agrawal, and Andrew McCallum.
\newblock Extending multi-sense word embedding to phrases and sentences for
  unsupervised semantic applications.
\newblock \emph{arXiv preprint arXiv:2103.15330}, 2021.

\bibitem[Choubey and Huang(2017)]{choubey-huang-2017-event}
Prafulla~Kumar Choubey and Ruihong Huang.
\newblock Event coreference resolution by iteratively unfolding
  inter-dependencies among events.
\newblock In \emph{Proceedings of the 2017 Conference on Empirical Methods in
  Natural Language Processing}, pages 2124--2133, Copenhagen, Denmark,
  September 2017. Association for Computational Linguistics.
\newblock \doi{10.18653/v1/D17-1226}.
\newblock URL \url{https://www.aclweb.org/anthology/D17-1226}.

\bibitem[Cybulska and Vossen(2014)]{cybulska2014using}
Agata Cybulska and Piek Vossen.
\newblock Using a sledgehammer to crack a nut? {L}exical {D}iversity and event
  coreference resolution.
\newblock In \emph{LREC}, pages 4545--4552, 2014.

\bibitem[Dhingra et~al.(2018)Dhingra, Jin, Yang, Cohen, and
  Salakhutdinov]{dhingra-etal-2018-neural}
Bhuwan Dhingra, Qiao Jin, Zhilin Yang, William Cohen, and Ruslan Salakhutdinov.
\newblock Neural models for reasoning over multiple mentions using coreference.
\newblock In \emph{Proceedings of the 2018 Conference of the North {A}merican
  Chapter of the Association for Computational Linguistics: Human Language
  Technologies, Volume 2 (Short Papers)}, pages 42--48, New Orleans, Louisiana,
  June 2018. Association for Computational Linguistics.
\newblock \doi{10.18653/v1/N18-2007}.
\newblock URL \url{https://www.aclweb.org/anthology/N18-2007}.

\bibitem[Eirew et~al.(2021)Eirew, Cattan, and Dagan]{Eirew2021WEC}
Alon Eirew, Arie Cattan, and Ido Dagan.
\newblock {WEC}: {D}eriving a {L}arge-scale {C}ross-document {E}vent
  {C}oreference dataset from {W}ikipedia.
\newblock In \emph{NAACL}. Association for Computational Linguistics, June
  2021.

\bibitem[{Falcon et al.}(2019)]{falcon2019pytorch}
William {Falcon et al.}
\newblock Pytorch lightning.
\newblock \emph{GitHub. Note:
  https://github.com/PyTorchLightning/pytorch-lightning}, 3, 2019.

\bibitem[Falke et~al.(2017)Falke, Meyer, and Gurevych]{falke-etal-2017-concept}
Tobias Falke, Christian~M. Meyer, and Iryna Gurevych.
\newblock Concept-map-based multi-document summarization using concept
  coreference resolution and global importance optimization.
\newblock In \emph{Proceedings of the Eighth International Joint Conference on
  Natural Language Processing (Volume 1: Long Papers)}, pages 801--811, Taipei,
  Taiwan, November 2017. Asian Federation of Natural Language Processing.
\newblock URL \url{https://www.aclweb.org/anthology/I17-1081}.

\bibitem[Glockner et~al.(2018)Glockner, Shwartz, and
  Goldberg]{glockner-etal-2018-breaking}
Max Glockner, Vered Shwartz, and Yoav Goldberg.
\newblock Breaking {NLI} systems with sentences that require simple lexical
  inferences.
\newblock In \emph{Proceedings of the 56th Annual Meeting of the Association
  for Computational Linguistics (Volume 2: Short Papers)}, pages 650--655,
  Melbourne, Australia, July 2018. Association for Computational Linguistics.
\newblock \doi{10.18653/v1/P18-2103}.
\newblock URL \url{https://www.aclweb.org/anthology/P18-2103}.

\bibitem[Gururangan et~al.(2020)Gururangan, Marasovi{\'c}, Swayamdipta, Lo,
  Beltagy, Downey, and Smith]{gururangan-etal-2020-dont}
Suchin Gururangan, Ana Marasovi{\'c}, Swabha Swayamdipta, Kyle Lo, Iz~Beltagy,
  Doug Downey, and Noah~A. Smith.
\newblock Don{'}t stop pretraining: Adapt language models to domains and tasks.
\newblock In \emph{Proceedings of the 58th Annual Meeting of the Association
  for Computational Linguistics}, pages 8342--8360, Online, July 2020.
  Association for Computational Linguistics.
\newblock \doi{10.18653/v1/2020.acl-main.740}.
\newblock URL \url{https://www.aclweb.org/anthology/2020.acl-main.740}.

\bibitem[Hearst(1992)]{hearst1992automatic}
Marti~A Hearst.
\newblock Automatic acquisition of hyponyms from large text corpora.
\newblock In \emph{Coling 1992 volume 2: The 15th international conference on
  computational linguistics}, 1992.

\bibitem[Hope et~al.(2017)Hope, Chan, Kittur, and Shahaf]{hope2017accelerating}
Tom Hope, Joel Chan, Aniket Kittur, and Dafna Shahaf.
\newblock Accelerating innovation through analogy mining.
\newblock In \emph{Proceedings of the 23rd ACM SIGKDD International Conference
  on Knowledge Discovery and Data Mining}, pages 235--243, 2017.

\bibitem[Hope et~al.(2020)Hope, Portenoy, Vasan, Borchardt, Horvitz, Weld,
  Hearst, and West]{hope2020scisight}
Tom Hope, Jason Portenoy, Kishore Vasan, Jonathan Borchardt, Eric Horvitz,
  Daniel~S Weld, Marti~A Hearst, and Jevin West.
\newblock Scisight: Combining faceted navigation and research group detection
  for {COVID-19} exploratory scientific search.
\newblock In \emph{Proceedings of the 2020 Conference on Empirical Methods in
  Natural Language Processing: System Demonstrations}, pages 135--143, 2020.

\bibitem[Hope et~al.(2021)Hope, Amini, Wadden, van Zuylen, Parasa, Horvitz,
  Weld, Schwartz, and Hajishirzi]{hope2021mechanisms}
Tom Hope, Aida Amini, David Wadden, Madeleine van Zuylen, Sravanthi Parasa,
  Eric Horvitz, Daniel Weld, Roy Schwartz, and Hannaneh Hajishirzi.
\newblock {Extracting a Knowledge Base of Mechanisms from COVID-19 Papers }.
\newblock 2021.

\bibitem[Jain et~al.(2020)Jain, van Zuylen, Hajishirzi, and
  Beltagy]{jain-etal-2020-scirex}
Sarthak Jain, Madeleine van Zuylen, Hannaneh Hajishirzi, and Iz~Beltagy.
\newblock {S}ci{REX}: {A} challenge dataset for document-level information
  extraction.
\newblock In \emph{Proceedings of the 58th Annual Meeting of the Association
  for Computational Linguistics}, pages 7506--7516, Online, July 2020.
  Association for Computational Linguistics.
\newblock \doi{10.18653/v1/2020.acl-main.670}.
\newblock URL \url{https://www.aclweb.org/anthology/2020.acl-main.670}.

\bibitem[Johnson et~al.(2017)Johnson, Douze, and J{\'e}gou]{JDH17}
Jeff Johnson, Matthijs Douze, and Herv{\'e} J{\'e}gou.
\newblock Billion-scale similarity search with gpus.
\newblock \emph{arXiv preprint arXiv:1702.08734}, 2017.

\bibitem[Kenyon-Dean et~al.(2018)Kenyon-Dean, Cheung, and
  Precup]{kenyon-dean-etal-2018-resolving}
Kian Kenyon-Dean, Jackie Chi~Kit Cheung, and Doina Precup.
\newblock Resolving event coreference with supervised representation learning
  and clustering-oriented regularization.
\newblock In \emph{Proceedings of the Seventh Joint Conference on Lexical and
  Computational Semantics}, pages 1--10, New Orleans, Louisiana, June 2018.
  Association for Computational Linguistics.
\newblock \doi{10.18653/v1/S18-2001}.
\newblock URL \url{https://www.aclweb.org/anthology/S18-2001}.

\bibitem[Lee et~al.(1993)Lee, Kim, and Lee]{Lee1993InformationRB}
J.~Lee, M.~Kim, and Yoon-Joon Lee.
\newblock Information retrieval based on conceptual distance in is-a
  hierarchies.
\newblock \emph{J. Documentation}, 49:\penalty0 188--207, 1993.

\bibitem[Lee et~al.(2017)Lee, He, Lewis, and Zettlemoyer]{lee-etal-2017-end}
Kenton Lee, Luheng He, Mike Lewis, and Luke Zettlemoyer.
\newblock End-to-end neural coreference resolution.
\newblock In \emph{Proceedings of the 2017 Conference on Empirical Methods in
  Natural Language Processing}, pages 188--197, Copenhagen, Denmark, September
  2017. Association for Computational Linguistics.
\newblock \doi{10.18653/v1/D17-1018}.
\newblock URL \url{https://www.aclweb.org/anthology/D17-1018}.

\bibitem[Li et~al.(2018)Li, Vilnis, Zhang, Boratko, and
  McCallum]{li2018smoothing}
Xiang Li, Luke Vilnis, Dongxu Zhang, Michael Boratko, and Andrew McCallum.
\newblock Smoothing the geometry of probabilistic box embeddings.
\newblock In \emph{International Conference on Learning Representations}, 2018.

\bibitem[Lin et~al.(2012)Lin, Etzioni, et~al.]{lin2012no}
Thomas Lin, Oren Etzioni, et~al.
\newblock No noun phrase left behind: {D}etecting and typing unlinkable
  entities.
\newblock In \emph{Proceedings of the 2012 joint conference on empirical
  methods in natural language processing and computational natural language
  learning}, pages 893--903, 2012.

\bibitem[Liu et~al.(2019)Liu, Ott, Goyal, Du, Joshi, Chen, Levy, Lewis,
  Zettlemoyer, and Stoyanov]{liu2019roberta}
Yinhan Liu, Myle Ott, Naman Goyal, Jingfei Du, Mandar Joshi, Danqi Chen, Omer
  Levy, Mike Lewis, Luke Zettlemoyer, and Veselin Stoyanov.
\newblock Ro{BERT}a: A robustly optimized {BERT} pretraining approach.
\newblock \emph{arXiv preprint arXiv:1907.11692}, 2019.

\bibitem[Lo et~al.(2020)Lo, Wang, Neumann, Kinney, and
  Weld]{lo-etal-2020-s2orc}
Kyle Lo, Lucy~Lu Wang, Mark Neumann, Rodney Kinney, and Daniel Weld.
\newblock {S}2{ORC}: The semantic scholar open research corpus.
\newblock In \emph{Proceedings of the 58th Annual Meeting of the Association
  for Computational Linguistics}, pages 4969--4983, Online, July 2020.
  Association for Computational Linguistics.
\newblock \doi{10.18653/v1/2020.acl-main.447}.
\newblock URL \url{https://www.aclweb.org/anthology/2020.acl-main.447}.

\bibitem[Luan et~al.(2018)Luan, He, Ostendorf, and
  Hajishirzi]{luan-etal-2018-multi}
Yi~Luan, Luheng He, Mari Ostendorf, and Hannaneh Hajishirzi.
\newblock Multi-task identification of entities, relations, and coreference for
  scientific knowledge graph construction.
\newblock In \emph{Proceedings of the 2018 Conference on Empirical Methods in
  Natural Language Processing}, pages 3219--3232, Brussels, Belgium,
  October-November 2018. Association for Computational Linguistics.
\newblock \doi{10.18653/v1/D18-1360}.
\newblock URL \url{https://www.aclweb.org/anthology/D18-1360}.

\bibitem[Luo(2005)]{luo-2005-coreference}
Xiaoqiang Luo.
\newblock On coreference resolution performance metrics.
\newblock In \emph{Proceedings of Human Language Technology Conference and
  Conference on Empirical Methods in Natural Language Processing}, pages
  25--32, Vancouver, British Columbia, Canada, October 2005. Association for
  Computational Linguistics.
\newblock URL \url{https://www.aclweb.org/anthology/H05-1004}.

\bibitem[Moosavi and Strube(2016)]{moosavi-strube-2016-coreference}
Nafise~Sadat Moosavi and Michael Strube.
\newblock Which coreference evaluation metric do you trust? a proposal for a
  link-based entity aware metric.
\newblock In \emph{Proceedings of the 54th Annual Meeting of the Association
  for Computational Linguistics (Volume 1: Long Papers)}, pages 632--642,
  Berlin, Germany, August 2016. Association for Computational Linguistics.
\newblock \doi{10.18653/v1/P16-1060}.
\newblock URL \url{https://www.aclweb.org/anthology/P16-1060}.

\bibitem[Mysore et~al.(2021)Mysore, O'Gorman, McCallum, and
  Zamani]{mysore2021csfcube}
Sheshera Mysore, Tim O'Gorman, Andrew McCallum, and Hamed Zamani.
\newblock {CSFC}ube--{A} test collection of computer science research articles
  for faceted query by example.
\newblock \emph{arXiv preprint arXiv:2103.12906}, 2021.

\bibitem[Paszke et~al.(2019)Paszke, Gross, Massa, Lerer, Bradbury, Chanan,
  Killeen, Lin, Gimelshein, Antiga, Desmaison, Kopf, Yang, DeVito, Raison,
  Tejani, Chilamkurthy, Steiner, Fang, Bai, and Chintala]{NEURIPS2019_9015}
Adam Paszke, Sam Gross, Francisco Massa, Adam Lerer, James Bradbury, Gregory
  Chanan, Trevor Killeen, Zeming Lin, Natalia Gimelshein, Luca Antiga, Alban
  Desmaison, Andreas Kopf, Edward Yang, Zachary DeVito, Martin Raison, Alykhan
  Tejani, Sasank Chilamkurthy, Benoit Steiner, Lu~Fang, Junjie Bai, and Soumith
  Chintala.
\newblock Pytorch: An imperative style, high-performance deep learning library.
\newblock In H.~Wallach, H.~Larochelle, A.~Beygelzimer, F.~d\textquotesingle
  Alch\'{e}-Buc, E.~Fox, and R.~Garnett, editors, \emph{Advances in Neural
  Information Processing Systems 32}, pages 8024--8035. Curran Associates,
  Inc., 2019.
\newblock URL
  \url{http://papers.neurips.cc/paper/9015-pytorch-an-imperative-style-high-performance-deep-learning-library.pdf}.

\bibitem[Poon and Domingos(2010)]{poon2010unsupervised}
Hoifung Poon and Pedro Domingos.
\newblock Unsupervised ontology induction from text.
\newblock In \emph{Proceedings of the 48th annual meeting of the Association
  for Computational Linguistics}, pages 296--305, 2010.

\bibitem[Pradhan et~al.(2012)Pradhan, Moschitti, Xue, Uryupina, and
  Zhang]{pradhan-etal-2012-conll}
Sameer Pradhan, Alessandro Moschitti, Nianwen Xue, Olga Uryupina, and Yuchen
  Zhang.
\newblock {C}o{NLL}-2012 shared task: Modeling multilingual unrestricted
  coreference in {O}nto{N}otes.
\newblock In \emph{Joint Conference on {EMNLP} and {C}o{NLL} - Shared Task},
  pages 1--40, Jeju Island, Korea, July 2012. Association for Computational
  Linguistics.
\newblock URL \url{https://www.aclweb.org/anthology/W12-4501}.

\bibitem[Ravenscroft et~al.(2021)Ravenscroft, Cattan, Clare, Dagan, and
  Liakata]{Ravenscroft2021CD2CRCR}
James Ravenscroft, Arie Cattan, Amanda Clare, Ido Dagan, and Maria Liakata.
\newblock {CD2CR}: Co-reference resolution across documents and domains.
\newblock In \emph{Proceedings of the 16th Conference of the {E}uropean Chapter
  of the Association for Computational Linguistics: Volume 1, Long Papers}.
  Association for Computational Linguistics, 2021.

\bibitem[Reimers and Gurevych(2019)]{reimers-2019-sentence-bert}
Nils Reimers and Iryna Gurevych.
\newblock Sentence-{BERT}: {S}entence embeddings using siamese bert-networks.
\newblock In \emph{Proceedings of the 2019 Conference on Empirical Methods in
  Natural Language Processing}. Association for Computational Linguistics, 11
  2019.
\newblock URL \url{https://arxiv.org/abs/1908.10084}.

\bibitem[Roller et~al.(2018)Roller, Kiela, and Nickel]{roller2018hearst}
Stephen Roller, Douwe Kiela, and Maximilian Nickel.
\newblock Hearst patterns revisited: Automatic hypernym detection from large
  text corpora.
\newblock In \emph{Proceedings of the 56th Annual Meeting of the Association
  for Computational Linguistics (Volume 2: Short Papers)}, pages 358--363,
  2018.

\bibitem[Shang et~al.(2020)Shang, Zhang, Liu, Li, and Han]{shang2020nettaxo}
Jingbo Shang, Xinyang Zhang, Liyuan Liu, Sha Li, and Jiawei Han.
\newblock Nettaxo: Automated topic taxonomy construction from text-rich
  network.
\newblock In \emph{Proceedings of The Web Conference 2020}, pages 1908--1919,
  2020.

\bibitem[Shwartz and Dagan(2016)]{shwartz-dagan-2016-adding}
Vered Shwartz and Ido Dagan.
\newblock Adding context to semantic data-driven paraphrasing.
\newblock In \emph{Proceedings of the Fifth Joint Conference on Lexical and
  Computational Semantics}, pages 108--113, Berlin, Germany, August 2016.
  Association for Computational Linguistics.
\newblock \doi{10.18653/v1/S16-2013}.
\newblock URL \url{https://www.aclweb.org/anthology/S16-2013}.

\bibitem[Shwartz et~al.(2016)Shwartz, Goldberg, and
  Dagan]{shwartz2016improving}
Vered Shwartz, Yoav Goldberg, and Ido Dagan.
\newblock Improving hypernymy detection with an integrated path-based and
  distributional method.
\newblock In \emph{Proceedings of the 54th Annual Meeting of the Association
  for Computational Linguistics (Volume 1: Long Papers)}, pages 2389--2398,
  2016.

\bibitem[Vilain et~al.(1995)Vilain, Burger, Aberdeen, Connolly, and
  Hirschman]{vilain-etal-1995-model}
Marc Vilain, John Burger, John Aberdeen, Dennis Connolly, and Lynette
  Hirschman.
\newblock A model-theoretic coreference scoring scheme.
\newblock In \emph{Sixth Message Understanding Conference ({MUC}-6):
  Proceedings of a Conference Held in {C}olumbia, {M}aryland, November 6-8,
  1995}, 1995.
\newblock URL \url{https://www.aclweb.org/anthology/M95-1005}.

\bibitem[Vossen et~al.(2018)Vossen, Ilievski, Postma, and
  Segers]{vossen-etal-2018-dont}
Piek Vossen, Filip Ilievski, Marten Postma, and Roxane Segers.
\newblock Don{'}t annotate, but validate: a data-to-text method for capturing
  event data.
\newblock In \emph{Proceedings of the Eleventh International Conference on
  Language Resources and Evaluation ({LREC} 2018)}, Miyazaki, Japan, May 2018.
  European Language Resources Association (ELRA).
\newblock URL \url{https://www.aclweb.org/anthology/L18-1480}.

\bibitem[Wadden et~al.(2019)Wadden, Wennberg, Luan, and
  Hajishirzi]{wadden-etal-2019-entity}
David Wadden, Ulme Wennberg, Yi~Luan, and Hannaneh Hajishirzi.
\newblock Entity, relation, and event extraction with contextualized span
  representations.
\newblock In \emph{Proceedings of the 2019 Conference on Empirical Methods in
  Natural Language Processing and the 9th International Joint Conference on
  Natural Language Processing (EMNLP-IJCNLP)}, pages 5784--5789, Hong Kong,
  China, November 2019. Association for Computational Linguistics.
\newblock \doi{10.18653/v1/D19-1585}.
\newblock URL \url{https://www.aclweb.org/anthology/D19-1585}.

\bibitem[Wang et~al.(2019)Wang, Yu, Guo, Das, Xiong, and
  Gao]{wang-etal-2019-multi-hop}
Haoyu Wang, Mo~Yu, Xiaoxiao Guo, Rajarshi Das, Wenhan Xiong, and Tian Gao.
\newblock Do multi-hop readers dream of reasoning chains?
\newblock In \emph{Proceedings of the 2nd Workshop on Machine Reading for
  Question Answering}, pages 91--97, Hong Kong, China, November 2019.
  Association for Computational Linguistics.
\newblock \doi{10.18653/v1/D19-5813}.
\newblock URL \url{https://www.aclweb.org/anthology/D19-5813}.

\bibitem[Ward~Jr(1963)]{ward1963hierarchical}
Joe~H Ward~Jr.
\newblock Hierarchical {G}rouping to {O}ptimize an {O}bjective {F}unction.
\newblock \emph{Journal of the American statistical association}, 1963.

\bibitem[Wolf et~al.(2020)Wolf, Debut, Sanh, Chaumond, Delangue, Moi, Cistac,
  Rault, Louf, Funtowicz, Davison, Shleifer, von Platen, Ma, Jernite, Plu, Xu,
  Le~Scao, Gugger, Drame, Lhoest, and Rush]{wolf-etal-2020-transformers}
Thomas Wolf, Lysandre Debut, Victor Sanh, Julien Chaumond, Clement Delangue,
  Anthony Moi, Pierric Cistac, Tim Rault, Remi Louf, Morgan Funtowicz, Joe
  Davison, Sam Shleifer, Patrick von Platen, Clara Ma, Yacine Jernite, Julien
  Plu, Canwen Xu, Teven Le~Scao, Sylvain Gugger, Mariama Drame, Quentin Lhoest,
  and Alexander Rush.
\newblock Transformers: State-of-the-art natural language processing.
\newblock In \emph{Proceedings of the 2020 Conference on Empirical Methods in
  Natural Language Processing: System Demonstrations}, pages 38--45, Online,
  October 2020. Association for Computational Linguistics.
\newblock \doi{10.18653/v1/2020.emnlp-demos.6}.
\newblock URL \url{https://www.aclweb.org/anthology/2020.emnlp-demos.6}.

\bibitem[Wu et~al.(2016)Wu, Song, and Giles]{wu2016exploring}
Zhaohui Wu, Yang Song, and C~Giles.
\newblock Exploring multiple feature spaces for novel entity discovery.
\newblock In \emph{Proceedings of the AAAI Conference on Artificial
  Intelligence}, volume~30, 2016.

\bibitem[Zhang et~al.(2018)Zhang, Tao, Chen, Shen, Jiang, Sadler, Vanni, and
  Han]{zhang2018taxogen}
Chao Zhang, Fangbo Tao, Xiusi Chen, Jiaming Shen, Meng Jiang, Brian Sadler,
  Michelle Vanni, and Jiawei Han.
\newblock Taxogen: Unsupervised topic taxonomy construction by adaptive term
  embedding and clustering.
\newblock In \emph{Proceedings of the 24th ACM SIGKDD International Conference
  on Knowledge Discovery \& Data Mining}, pages 2701--2709, 2018.

\end{thebibliography}
\bibliographystyle{plainnat}

\clearpage
\appendix
\maketitle
\section{Data Collection}
\label{app:data}

\subsection{Hypernym extraction details}
\label{app:hypernym}

As mentioned in the paper (§\ref{subsec:docsandmentions}), we build a large hypernym graph based on the extracted hypernym relations at the sentence level. For example, given the sentences (from two different papers): \emph{``The Travelling Salesman Problem (TSP) is one of the major problems in graph theory.''} and \emph{``Graph isomorphism is a major problem in graph theory.''}, the Dygie++ model identified the relations graph theory $\xrightarrow[]{}$ Travelling Salesman Problem (TSP) and graph theory $\xrightarrow[]{}$ Graph isomorphism.
In our corpus-level hypernym extraction procedure, we use standard surface-level string normalization to unify mentions across the corpus (removing punctuation, lemmatizing, lowercasing and the Levenshtein edit-distance with threshold 0.8), resulting in distinct 250K tasks and 1.2M methods with 340K and 1.4M hierarchical edges between them respectively. We then sample parent-child mentions and siblings to form our candidates (see Figure~\ref{fig:anno_proc} (1-2)).

In addition to surfacing candidates for the referential hierarchy, another significant advantage of this resource is indeed one of its main weaknesses (discussed in §\ref{sec:bg}): it produces coreferring mentions of the same concept in the form of hyponyms of a shared parent, which we use to enrich \data{} for the CDCR task (e.g., two different papers mentioning \textsl{image synthesis} or \textsl{image generation} as children of \textsl{computer vision tasks}).

While this automatically extracted hierarchy has broad coverage, it suffers from noise.\footnote{For example, from the text \emph{... image information for analysis purposes, such as segmentation, identification ...}, we obtain the hypernym relation (analysis purposes, identification). } To reduce noise, an annotator who did not take part in the final data annotation (§\ref{subsec:anno}) was given a set of generated candidates and post-processed to filter overly generic or noisy spans.

\subsection{Mention Retrieval Details}
\label{app:mention_retrieval}

Formally, denote by $\mathcal{S} =(V,E)$ the union of the PwC and hypernym graphs. For a parent concept $\mathrm{p} \in V$, let $\{\mathrm{c}\}_\mathrm{p}$ be its set of direct descendants (children). We take the parent vertex $\mathrm{p}$ and its child vertices $\{\mathrm{c}\}_\mathrm{p}$, and add the 60 curated groups to this set, to form a complete set of candidates we denote by $\mathrm{C}_\mathrm{p}$. 

For the sources making up $\mathcal{S}$, the concept names in PwC and the curated lists are detached from any specific paper context, which is required for building \data{}. While the extracted hypernym relations do come from specific sentences, using only those contexts as candidates would bias models to focus only on within-document relations. In addition, we wish to diversify our the surface forms of mentions by retrieving subtle variants (e.g., retrieving for \textsl{BERT} mentions such as \textsl{BERT model}, \textsl{BERT architecture}, \textsl{BERT-based representation}, etc.).

Thus, we augment each group by retrieving similar mentions from our corpus.  Specifically, for each $\mathrm{C}_\mathrm{p}$\, we find similar mentions for each $c \in \mathrm{C}_\mathrm{p}$. We use an encoding function $f:c\mapsto\mathbb{R}^d$ that maps the surface form of each selected $c$ to $f(c)$, a $d$ dimensional vector.\footnote{We also experimented with using the context of the mentions as well, but found this to result in more easy near-exact matches along with more highly noisy ones.}  Following the approach of \citep{reimers-2019-sentence-bert}, the encoding function is obtained by fine-tuning a language model pre-trained on computer science papers~\cite{gururangan-etal-2020-dont} on two semantic similarity tasks, STS \cite{cer2017semeval} and SNLI \cite{bowman2015large}. We then apply $f(\cdot)$ to all distinct mentions in our large-scale corpus with over 30M mentions. Finally, we augment each group with the top $K$ highest-scoring mentions by cosine similarity to each $c \in \mathrm{C}_\mathrm{p}$ in the initial group and take the union of retrieved results. To manage the scale, we employ a system designed for fast similarity-based search \cite{JDH17}.\footnote{We filter for mentions with similarity greater than $.8$, empirically observing sufficient diversity and precision.} To make sure we sample enough mentions from SciREX despite its comparatively small size, we sample from it with the same proportion as from the 10M abstracts.
%\footnote{In practice, many concepts in $\mathcal{S}$ are not present in SciREX (which is focused on AI subfields); we obtain 21297 mentions from the broader corpus and 5468 from SciREX.}

\subsection{Annotation Interface}
\label{app:annotation}

To annotate \data{}, we extend CoRefi~\citep{bornstein-etal-2020-corefi} with the ability to annotate hierarchy of clusters.\footnote{\url{https://github.com/ariecattan/CoRefi}} The hierarchy is kept in sync with the clusters -- any modification of the clusters (e.g., merging two clusters) affects automatically the hierarchy (e.g., unifying their children). Therefore, annotators can annotate both the clusters and the hierarchy at the same time or alternate between them. In addition, annotators can (and are encouraged to) add some notes in the hierarchy to help them distinguish between ambiguous concepts. As shown in Figure~\ref{fig:anno_proc}, we extend CoRefi by displaying metadata for each paper, including the link to the Semantic Scholar URL of the paper. This additional context is often helpful in order to annotate complex cases. For example, as shown in~Table~\ref{tab:phenomena}, for resolving the ambiguity between two mentions of \emph{cross-view training}, it may be useful to see that one paper was published a few years before the other one coined its own \emph{cross-view training} as a name for their semi-supervised model. 

%In §\ref{subsec:anno} we described a specific example of using a query by example setting to guide annotator. Here we present the example in its more general form. Consider two mentions $m_1$ and $m_2$. Assume we treat $m_1$ a search query. We retrieve a list of mentions in context, one of them being $m_2$. Also consider the inverse, where $m_2$ is the query and $m_1$ is returned.  
%If both directions (searching for $m_1$ and getting $m_2$, or searching for $m_2$ and getting $m_1$) are relevant, then both mentions belong in the same cluster due to either referring to the same concept (or a subtle variation around it); otherwise, if only a single direction is correct, then one should be the parent of the other; if both directions are considered incorrect, the mentions should be considered unrelated.

\section{Models}

\subsection{\citet{Cattan2021CrossdocumentCR}'s CDCR model}
\label{app:cattan_model}

Here, we describe the cross-document resolver of \cite{Cattan2021CrossdocumentCR} that we use as baseline for \data{} (§\ref{subsec:baselines}).
This model is based on contextualized token and span representations. For a document $d$, we are given a sequence of tokens $\mathbf{x} = \{x_1,x_2,\ldots,x_T\}$ where $T$ is the length of the document. We first obtain a contextualized embedding of each $x_t \in \mathbf{x}$ using the RoBERTa-large pre-trained language model~\citep{liu2019roberta}. Each mention span $i$ is a contiguous subsequence of $\mathbf{x}$, denoted $\mathbf{x}_i = \{x^{(i)}_1,x^{(i)}_2,\ldots,x^{(i)}_S\}$. Let $\mathbf{\hat{x}}_i = \{\hat{x}^{(i)}_1,\hat{x}^{(i)}_2,\ldots,\hat{x}^{(i)}_S\}$ be the sequence of embedded tokens in span $i$, where $\hat{x}^{(i)}_t$ is the embedding of token $t$ in span $i$. Then, similarly to~\citet{lee-etal-2017-end}, each span $i$ is represented (\ref{eq:repr}) by the concatenation of embeddings of the boundary tokens in the span ($\hat{x}^{(i)}_1$ and $\hat{x}^{(i)}_S$), an attention-weighted sum of the token embeddings in span $i$ (Attn$(\mathbf{\hat{x}}_i)$), and a feature vector denoting the span width (number of tokens).
\begin{equation}
\label{eq:repr}
    g_{i} = [\hat{x}^{(i)}_1, \hat{x}^{(i)}_S, \text{Attn}(\mathbf{\hat{x}}_i), \phi(i)] 
\end{equation}

Given a pair of contextualized span embeddings $g_i$ and $g_j$ from two different documents, we feed them to a pairwise scorer (\ref{eq:pairwise}) in the form of a simple feed-forward network that receives as input the concatenation of two span representations and their element-wise product, and outputs the likelihood that the two mentions corefer.
\begin{equation}
\label{eq:pairwise}
    s(i, j) = \text{FFNN}([g_i, g_j, g_i \circ g_j]) \\
\end{equation}

At inference time, the coreference clusters are formed by applying agglomerative clustering \cite{ward1963hierarchical} over a pairwise distance matrix populated with scores $s(i, j)$, using the average-linkage cluster merging criterion.  Following \citep{Cattan2021CrossdocumentCR}, this model does not involve fine-tuning of the underlying language model due to prohibitive memory costs --- it begins with encoding all documents using an LM, then computes the pairwise scores over all pairs of mention representations. We note that our unified approach is the only model we evaluate that fine-tunes the Longformer on our data. 

% \subsection{Longformer and CDLM}
% \label{app:longform}

% During pretraining, Longformer and CDLM apply \emph{local attention} --- attention only to tokens in a fixed-sized window around each token. When fine-tuning on a specific task, \emph{global attention} --- attention to all tokens in the sequence --- can be assigned to a few target tokens to encode global information.
% Following~\citet{Caciularu2021CrossDocumentLM}, we take each mention and its corresponding paragraph, inserting mention markers \texttt{<m>} and \texttt{</m>} surrounding the mention to obtain a mention representation. For CDLM, we add the document markers \texttt{<doc-s>} and \texttt{</doc-s>} surrounding each document. Then, we concatenate the representations of $m_1$ and $m_2$ separated by a separator token \texttt{</s>}, and add a \texttt{[CLS]} token at the beginning. We assign \emph{global attention} to the \texttt{[CLS]} and the mention markers of the two mentions. The \texttt{[CLS]} vector is finally fed into a linear layer $W^{d \times 4}$, for fine-tuning the model. 

\subsection{Hyper parameters for the Unified Model}
\label{app:hp}

We develop our model in Pytorch~\citep{NEURIPS2019_9015} and PytorchLightning~\citep{falcon2019pytorch} using the Transformers library~\cite{wolf-etal-2020-transformers} and the AdamW optimizer. We train our model on 1 epoch using a batch size of 4 and gradient accumulation of 4 and a learning rate of 1E-5. We conduct our experiments on 8 Tesla V100 32GB GPUS using distributed data parallelism. 
The training time takes about 2.5 hour for the unified model and inference 25 minutes. Our model includes 148M parameters. 
We fine-tune the threshold for the agglomerative clustering and the stopping criterion for the hierarchy on the validation set, in a range of \{0.4, 0.6\} for both, set to 0.6 and 0.4 respectively. We take the values that achieve the best path ratio metric on the validation set.

\section{Coreference and Hierarchy Results}
\label{app:coref_results}

Table~\ref{tab:hierarchy} presents the results of all models according to the cluster-level hierarchy score (§\ref{subsec:eval_metrics}) in terms of recall, precision and F1.

\begin{table*}[!t]
    \centering
     \resizebox{0.7\textwidth}{!}{
    \begin{tabular}{llllclll}
    \toprule
    & \multicolumn{3}{c}{Hierarchy} && \multicolumn{3}{c}{Hierarchy 50\%} \\
    Model & Recall & Precision & F1 && Recall & Precision & F1 \\
    \midrule
        CA\textsubscript{News} &  43.3 & 32.4 & 37.1 && 21.8 & 9.3 & 13.0 \\
        CA\textsubscript{Sci-News} & 37.8 & 23.7 & 29.2 && 12.2 & 12.5 & 12.3 \\
        \midrule
        CA\textsubscript{\data{}} & \textbf{45.5} & 16.0 & 23.7 && 17.7 & 14.3 & 15.8 \\
         CA\textsubscript{\data{}} + CS-RoBERTa & 43.6 & 13.1 & 23.5 && 19.5 & 13.8 & 16.1  \\
         CA\textsubscript{\data{}} + SciBERT & 41.4 & 16.7 & 23.8 && 27.2 & 13.3 & 17.8 \\
        %  Pipeline & 32.6 & 48.2 & 38.1 \\
        \midrule
        %  Unified\textsubscript{LF-sentence} & 33.8 & 59.0 & 43.0 && 25.4 & 50.8 & 33.9 \\
          Unified\textsubscript{Longformer} & 36.6 & 56.7 & 44.5 && \textbf{29.1} & 48.2 & \textbf{36.3} \\
          Unified\textsubscript{CDLM} & 36.4 & \textbf{58.2} & \textbf{44.8} && 27.8 & \textbf{49.1} & 35.5 \\
        \bottomrule
    \end{tabular}}
    \caption{Recall, Precision and F1 according to the Cluster-level Hierarchy Score (§\ref{subsec:eval_metrics}) for all models.}
    \label{tab:hierarchy}
\end{table*}

Table~\ref{tab:full_coref_results} presents the results of the inter-annotator agreement (IAA), all baseline models as well as the pipeline and unified, according to all common coreference metrics (MUC, B\textsuperscript{3}, CEAFe, LEA and CoNLL F1).
We obtain coreference metrics using the python implementation of the standard coreference metrics~\citep{moosavi-strube-2016-coreference}.\footnote{\url{https://github.com/ns-moosavi/coval/}} Following~\citet{cattan2021eval}, we apply coreference metrics only on non-singleton (gold and predicted) clusters in order to avoid inflated results.

\begin{table*}[!ht]
    \centering
    \resizebox{\textwidth}{!}{
    \begin{tabular}{@{}llccccccccccccccclc@{}}
    \toprule
    && \multicolumn{3}{c}{MUC} && \multicolumn{3}{c}{$B^3$} & & \multicolumn{3}{c}{$CEAFe$} && \multicolumn{3}{c}{LEA} && CoNLL\\
    \cmidrule{3-5} \cmidrule{7-9} \cmidrule{11-13} \cmidrule{15-17} \cmidrule{19-19}
    && R & P & $F_1$ && R & P & $F_1$ && R &P & $F_1$ && R &P & $F_1$ && $F_1$  \\ 
   \midrule
        IAA && - & - & 89.6 && - & - & 81.4 && - & - & 77.1 && - & - & 79.7 && 82.7 \\
       \midrule 
       CA\textsubscript{News} && 83.8 & 64.0 & 72.5 && 69.9 & 35.5 & 47.1  && 32.6 & 44.2 & 37.5 && 65.4 & 31.5 & 42.5 && 52.4 \\
        CA\textsubscript{Sci-News} && 75.4 & 65.5 & 70.1 && 66.5 & 24.8 & 36.1 && 17.4 & 41.4 & 24.5 && 63.6 & 22.0 & 32.7 && 43.5  \\
        \midrule 
        CA\textsubscript{\data{}} && 54.9 & 81.0 & 65.4 && 40.3 & 73.1 & 52.0 && 48.1 & 48.2 & 48.1 && 38.5 & 69.2 & 47.8 && 55.2 \\
         CA\textsubscript{\data{}} + CS-RoBERTa && 58.4 & 79.7 & 67.4 && 44.8 & 69.9 & 54.6 && 48.9 & 51.3 & 50.1 && 41.1 & 65.9 & 50.7 && 57.4 \\
         CA\textsubscript{\data{}} + SciBERT && 78.0 & 78.8 & 78.4 && 65.6 & 64.6 & 65.1 && 56.4 & 57.2 & 56.8 && 62.4 & 61.3 & 61.9 && 66.8 \\
         \midrule 
        %  Pipeline && 78.7 & 90.5 & 84.2 && 69.9 & 85.8 & 75.2 && 73.3 & 68.9 & 71.0 && 64.2 & 93.7 & 72.6 && 76.8 \\
        %  Unified\textsubscript{LF-sentence} && 86.2 & 85.9 & 86.0 && 76.0 & 75.3 & 75.6 && 69.5 & 68.9 & 69.2 && 73.6 & 72.7 & 73.2 && 77.0 \\
         Unified\textsubscript{Longformer} && \textbf{88.5} & 84.9 & \textbf{86.7} && \textbf{79.4} & 72.8 & \textbf{75.9} && 67.8 & \textbf{70.4} & \textbf{69.0} && \textbf{77.4} & 70.4 & \textbf{73.7} && \textbf{77.2} \\
         Unified\textsubscript{CDLM} && 87.4 & \textbf{85.3} & 86.3 && 77.8 & \textbf{73.8} & 75.8 && \textbf{68.0} & 70.0 & \textbf{69.0} && 75.6 & \textbf{71.3} & 73.4 && 77.0 \\
    \bottomrule
    \end{tabular}}
    \caption{Coreference results of the IAA, baseline models, pipeline and unified on the \data{} test set according to all coreference metrics: MUC, B\textsuperscript{3}, CEAFe, LEA and CoNLL F1. Since IAA is computed by considering one annotator as gold and the other as system, the IAA is reported only according to the symetric F1. }
    \label{tab:full_coref_results}
\end{table*}

\end{document}